\documentclass{article}

\usepackage[utf8]{inputenc}
\usepackage{graphicx}
\usepackage{amssymb}
\usepackage{epstopdf}
\usepackage{subcaption}
\usepackage{booktabs}
\usepackage{float}
\usepackage{hyperref}
\usepackage{threeparttable}
\usepackage{tabularx}
\usepackage{color}
\usepackage{todonotes}
\newcommand\runintitle[1]{\par\vspace{.5em}\noindent \textbf{#1}}
\newcommand{\plotslegend}{Legend: individual runs of {\color{red}$+$}~mGPTIPS, {\color{green}\textsf{Y}}~GPTIPS, {\color{cyan}\normalsize$\times$}~EFS, {\color{black}$\bullet$}~FFX, median RMSE of ---~LR, -~-~-~RF, $\cdots$~SVR.}
\newcommand{\keywords}[1]{\runintitle{Keywords} #1}
\newenvironment{acknowledgements}{\runintitle{Acknowledgements}}{}

\begin{document}

\title{Symbolic Regression Algorithms with Built-in Linear Regression\\
    {\normalsize A Comparison}
}

\author{Jan Žegklitz \and
        Petr Pošík}

\date{}

\maketitle

\begin{abstract}
    Recently, several algorithms for symbolic regression (SR) emerged which employ a form of multiple linear regression (LR) to produce generalized linear models.
    The use of LR allows the algorithms to create models with relatively small error right from the beginning of the search; such algorithms are thus claimed to be (sometimes by orders of magnitude) faster than SR algorithms based on vanilla genetic programming.
    However, a systematic comparison of these algorithms on a common set of problems is still missing.
    In this paper we conceptually and experimentally compare several representatives of such algorithms (GPTIPS, FFX, and EFS).
    They are applied as off-the-shelf, ready-to-use techniques, mostly using their default settings.
    The methods are compared on several synthetic and real-world SR benchmark problems.
    Their performance is also related to the performance of three conventional machine learning algorithms --- multiple regression, random forests and support vector regression.
\end{abstract}

\section{Introduction}
Symbolic regression (SR) is an inductive learning task with the goal to find a model in the form of a (preferably simple) symbolic mathematical expression that fits the available training data.
While the models produced by other well-known machine learning (ML) techniques for regression (e.g. neural networks, support vector machines, or random forests) are often useful, they are essentially black boxes which are hard to analyze.
On the other hand, SR aims to extract white box models, easy to analyze.

SR is a landmark application of Genetic Programming (GP) \cite{Koza1992}. GP is an evolutionary optimization technique that is inspired by biological evolution to evolve computer programs that perform well in a given task.
GP is similar to Genetic Algorithms \cite{Holland1992}: it uses a \emph{population} of individuals (candidate solutions), a \emph{fitness function} that evaluates the behavior of the solutions, a \emph{selection} mechanism to promote better solutions over the worse ones, a \emph{crossover} operator(s) that combines two (or more) individuals and a \emph{mutation} operator(s) that (randomly) modifies individuals.
The difference from GAs is that the evolved structure is not a fixed-sized array of binary or real numbers but a variable-sized data structure, typically a tree, that represents a \emph{program} that solves (or is supposed to solve) a given class of problems. Such a program can also be a mathematical expression. For the rest of this article we will refer to the Koza's original GP \cite{Koza1992} system as to `vanilla GP'.

When vanilla GP is applied to a SR task, it usually needs a relatively long time to find an acceptable solution.
While the conventional ML techniques usually fit models with a structure fixed in advance and only tune the parameters, GP searches a much broader class of possible models limited only by the user, usually by specifying the sets of function and terminal symbols, and maximal model complexity.
In other words, GP searches also for a useful structure of the model.
Such a system may reach impressive results \cite{Schmidt2009,eureqa} when given good data and enough time, sometimes even recovering the true equations describing the underlying phenomenon which generated the observed data.

A novel, revealing view of the SR problem is provided by Geometric Semantic Genetic Programming (GSGP) \cite{Moraglio2012}.
The authors put emphasis on the difference between syntax (the actual trees and expressions) and the semantics (the output values of the candidate functions).
The semantic space is $n$-dimensional euclidean space where $n$ is the number of test cases.
Each candidate function maps into this semantic space as a single point, with coordinates equal to the errors the function makes for individual test cases.
From this point of view, the goal is to find a function that lies as close as possible to the origin of the semantic space.

GSGP uses simple linear operators to search the semantic space.
Crossover takes two trees from the population and creates an offspring by constructing a tree representing a (weighted) average of the parents.
Mutation takes an individual and produces an offspring by linear combination of the parent and a randomly generated tree (which is itself generated as a difference of 2 random trees).
From the point of view of these operators, the fitness landscape is unimodal, hence easy to search.
GSGP is able to converge very quickly (compared to vanilla GP) and steadily. It is also resistant to overfitting thanks to the small steps it is taking towards the optimum.
On the other hand, GSGP's major disadvantage is the fact that the size of a solution grows exponentially with time, resulting in huge trees, that are (i) effectively black-box and (ii) slow to evaluate (even though this can be alleviated by a careful housekeeping).

A combination of GSGP with Local Search \cite{Castelli2015} proposed recently
uses only the mutation, but the offspring is constructed as the \emph{optimal} linear combination with respect to the parent and a random tree via multiple regression.
Using only this local search operator, the GSGP-LS converges much faster than GSGP on the training sets (though, it is also much more susceptible to overfitting).
In the end, however, both the above mentioned versions of GSGP produce models which have the form of a linear combination of randomly generated trees.

Recently, several methods emerged \cite{Searson2010,Searson2015-published,McConaghy2011,Arnaldo2015,Arnaldo2014} that explicitly restrict the class of models to generalized linear models, i.e. to a linear combination of possibly non-linear basis functions.
With the help of linear regression techniques applied to the basis functions, such models can be learned much faster.
In \cite{McConaghy2011}, it is argued that (some of) these SR methods already have the status of \emph{a~technology}, i.e. that they are available to their prospective users as off-the-shelf, ready-to-use tools that can be simply applied to available data, without modifying their internals or investing much effort to tune the method.

The first goal of this paper is to evaluate and compare several recent algorithms for SR which---according to our opinion---are close to being a technology.
We chose 3 methods\footnote{
    Another candidate for such a comparison would be system Eureqa \cite{Schmidt2009,eureqa}.
    However, we decided not to include it in the comparison because (1) it is currently a commercial software and we want to focus on open-source solutions freely available to anyone, and (2) the free academic licence does not contain any API for automating the system.
    We also exclude GSGP systems, since they tend to create too complex models, and the model creation process does not contain an explicit use of multiple linear regression on the global level.
}: (1) GPTIPS \cite{Searson2015-published}, a SR framework using multigene genetic programming, (2) Fast Function Extraction (FFX) \cite{McConaghy2011}, an example of non-evolutionary deterministic methods, and (3) Evolutionary Feature Synthesis (EFS) \cite{Arnaldo2015}, a recent evolutionary method for fast creation of interpretable SR models.
All these methods were reported by their authors to be successful SR solvers creating simple and interpretable models.
We do not expect that one of the above algorithms would produce better models in all reasonable circumstances (cf. No Free Lunch theorems for supervised learning \cite{Wolpert2001});
we are more interested in the types of differences we can expect from these algorithms when applied to the same regression problems. 
To the best of our knowledge, such a comparison has not been done yet.

In this paper, all the above algorithms are used with their default parameter settings (or with minimal changes allowing a reasonable comparison).
They are applied to 5 synthetic and 4 real-world SR problems of varying complexity.
The synthetic problems contain internal constants\footnote{
    By an internal constant we mean a constant other than a coefficient of a top-level linear combination.
    Example: in $3x^2 + 6\sin(1.3x)$, the ,,3`` and ,,6`` are not internal constants, because these are tuned by the top-level multiple regression, while the ,,1.3`` is internal constant (part of the nonlinear basis function).
} which are hard to find for all these algorithms.
The results on the real-world problems should show whether the inability to find internal constants prevents the methods from finding a useful model.

The second goal of this article is to provide a meaningfull baselines for the comparison of the above SR methods. 
For the classical ML methods it is nowadays common to tune their hyperparameters; from our point of view this is also a ready-to-use \emph{technology}.
It is thus fair to include in the comparison a few baselines constituted by conventional ML methods (pure multiple regression, random forest, and support vector regression) with their hyperparameters tuned using a grid search
(as opposed to comparing with ML methods with fixed, arbitrarily chosen hyperparameters as done in the original articles).
This way we will compare SR methods which optimize the model expression structure within the given model complexity constraints (and with a very limited ability to tune the internal constants of the models) on the one hand, and on the other hand ML methods which use fixed-structure models with varying complexity (set by the grid search over hyperparameters), which are able to tune their internal constants very well.

The rest of the article is organized as follows: in Section \ref{sec:algorithms}, the compared algorithms are described in more detail.
Section \ref{sec:benchmarks} then introduces the benchmark problems we use to compare the SR methods, and also describes the experimental methodology.
Section \ref{sec:results} contains the results and their discussion.
Section \ref{sec:conclusion} concludes the paper and provides suggestions for future work.

\section{Compared Algorithms} \label{sec:algorithms}
This section briefly describes the selected algorithms and important aspects regarding the complexity of models produced by these algorithms.

\subsection{GPTIPS}
GPTIPS \cite{Searson2010,Searson2015-published} is an open-source SR toolbox for MATLAB.
It is an implementation of Multi-Gene Genetic Programming (MGGP) \cite{Hinchliffe1996} and thus has its roots in vanilla GP.
Each solution is composed of multiple independent trees, called genes, and their outputs are linearly combined.
The coefficients of this linear combination are computed optimally with respect to the mean squared error (MSE) of the resulting expression measured on the training data using ordinary least squares method.

MGGP (and GPTIPS in particular) is based on classical Genetic Programming.
This means that it works with a population of fixed size, subtree mutation, subtree crossover, tournament selection, standard initialization procedures, and is able to handle the internal constants of the model (to certain extent) using ephemeral random constants.
The output of GPTIPS is the last population of models (not a pareto front); it is up to the user to choose the final one.

To limit the complexity of the candidate models and to prefer simpler ones, GPTIPS by default uses Lexicographic Parsimony Pressure \cite{Luke2002} using Expressional Complexity \cite{Vladislavleva2009} of the models (genes).
The top-level linear combination of the models is not restricted (regularized) in any way.

MGGP was shown to be faster and more accurate than vanilla GP \cite{Hinchliffe1996} and also a comparable or better alternative to classical methods like Support Vector Regression and Artificial Neural Networks \cite{Garg2013}.

\subsection{FFX}
FFX, or Fast Function Extraction \cite{McConaghy2011}, is a \emph{deterministic} algorithm for symbolic regression.
It first exhaustively generates a massive set of basis functions, which are then linearly combined using Pathwise Regularized Learning \cite{Friedman2010,Zhou2005} to produce sparse models.
The algorithm produces a pareto-front of models with respect to their accuracy and complexity.
Again, it is up to the user to choose the final model.

There are two kinds of bases that are generated: univariate bases and bivariate bases.
Univariate bases are: a variable raised to a power (chosen from a fixed set of options) and (non-linear) functions applied to another univariate base.
Bivariate bases are products of all pairs of univariate bases excluding the pairs where both the bases are of function-type; the author argues that such products are ``deemed to be too complex.'' 
FFX also includes a trick that allows it to produce rational functions of the bases using the same learning procedure.

The original paper \cite{McConaghy2011} reports FFX to be more accurate than many classical methods including vanilla GP, neural networks and SVM.

\subsection{EFS}
EFS, or Evolutionary Feature Synthesis \cite{Arnaldo2015}, is the most recent of the three algorithms.
In EFS, the population does not consist of complete models but rather of features which, collectively, form a single model.
In this respect EFS is similar to FFX: in FFX the individual features are relatively simple and are generated systematically and exhaustively, while in EFS, features may be more complex (depending on the complexity constraints) and are generated stochastically.

The initial population is formed by the original features of the dataset.
Then, in each generation, a model is composed of the features in the current population by Pathwise Regularized Learning and is stored if it is the best.
The next step in a generation is the composition of new features by applying unary and binary functions to the features already present in the current population.
This way, more complex features are created from simpler ones.
Also, the features are selected during this composition step according to the Pearson correlation coefficient with the feature's parents.

EFS does not build the symbolic model explicitly -- it works with the data of the features in a vectorial fashion and only stores the structure for logging purposes.
This results in a very fast algorithm.

The original paper \cite{Arnaldo2015} reports EFS being comparable to neural networks and similar or better than Multiple Regression Genetic Programming which itself was reported to outperform vanilla GP, multiple regression and Scaled Symbolic Regression (introduced in \cite{Keijzer2004}).

\subsection{Model Complexity Constraints}
Each algorithm described above handles the issue of resulting model complexity in a different way.
GPTIPS has (user-defined) limits on the maximum number of nodes and/or maximum depth, and on the maximum number of bases.
By default there is a depth limit of 4, and maximum number of bases (not counting the intercept) is also 4.
EFS computes the maximum number of bases from the number of input features; maximum number of nodes in a base is hard-coded to 5.
The FFX procedure results in a maximum model depth of 5.

\section{Benchmarks and Testing} \label{sec:benchmarks}
For testing, we selected five artificial and four real-world benchmarks.
The artificial benchmarks cover various types of complexities and features.
An important feature of all the artificial benchmarks except Koza-1 is that they contain internal constants, which is challenging for all the algorithms.
In case of the real-world benchmarks, the ground truth, i.e. the function that generated the data, is not known.
The quality of the results is judged just by the testing error: we shall thus see whether the inability to learn the internal constants is a show-stopper for these algorithms.

\subsection{Artificial Benchmarks}
All the datasets except the last one were picked based on \cite{McDermott2012}.
Table \ref{tab:datasets-artificial} presents a summary of the used artificial benchmarks: their definitions, number of dimensions and their original source.
Table \ref{tab:datasets-artificial-sampling} presents the training and testing sampling of those datasets. 
Using the notation from \cite{McDermott2012}:
\begin{itemize}
    \item the expression $U[a, b, c]$ means $c$ random samples uniformly distributed in the interval $[a, b]$ for each variable;
    \item the expression $E[a, b, c]$ means a grid in the interval $[a, b]$ with spacing of $c$ for each variable.
\end{itemize}

\begin{table}[ht]
    \centering
    \caption{Definitions of the artificial benchmarks.}
    \label{tab:datasets-artificial}
    \begin{tabular}{llcc}
        \toprule
        Name & Definition & Dim & Ref \\
        \midrule
        Koza-1
        & $f_1(x) = x^4 + x^3 + x^2 + x$
        & 1
        & \cite{Koza1992}
        \\
        Korns-11
        & $f_2(x, y, z, v, w) = 6.87 + 11 \cos(7.23x^3)$
        & 5
        & \cite{Korns2011}
        \\
        S1
        & $f_3(x) = \mathrm{e}^{-x}x^3\sin(x)\cos(x)(\sin^2(x)\cos(x)-1)$
        & 1
        & \cite{Vladislavleva2009}
        \\
        S2
        & $f_4(x, y) = (y-5)f_3(x)$
        & 2
        & \cite{Vladislavleva2009}
        \\
        UB
        & $f_5(x_1,x_2,x_3,x_4,x_5) = \frac{10}{5+\sum_{i=1}^5(x_i-3)^2}$
        & 5
        & \cite{Vladislavleva2009}
        \\
        \bottomrule
    \end{tabular}
\end{table}

\begin{table}[ht]
    \centering
    \caption{
        Description of the training and testing sampling. (Each variable in S2 has its own sampling type.)
    }
    \label{tab:datasets-artificial-sampling}
    \begin{tabular}{lll}
        \toprule
        Name & Training sampling & Testing sampling \\
        \midrule
        Koza-1
        & $U[-1, 1, 20]$
        & $U[-1, 1, 100]$
        \\
        Korns-11
        & $ U[-50, 10, 10000] $
        & $ U[-50, 10, 10000] $
        \\
        S1
        & $ E[-0.5, 10.5, 0.1] $
        & $ E[-0.5, 10.5, 0.05] $
        \\
        S2
        & $ x = E[-0.5, 10.5, 0.1] $
        & $ x = E[-0.5, 10.5, 0.05] $
        \\
        & $ y = E[-0.5, 10.5, 2] $
        & $ y = E[-0.5, 10.5, 0.5] $
        \\
        UB
        & $ U[-0.25, 6.35, 1024] $
        & $ U[-0.25, 6.35, 5000] $
        \\
        \bottomrule
    \end{tabular}
\end{table}

\runintitle{Koza-1} \cite{Koza1992}
is a classical, easy-to-solve SR benchmark.
It shall test the ability of the algorithms to fit a very simple function.

\runintitle{Korns-11} \cite{Korns2011}
is specific in the fact that the output depends on only one of the 5 input features and also by the presence of internal constant.
The function is hard to fit because of the high frequency components.

\runintitle{Salustowicz 1D (S1)} \cite{Vladislavleva2009}
(called Vladislavleva-2 in \cite{McDermott2012}) is defined by a single, relatively complex term.
It does not fit the generalized linear model structure well.

\runintitle{Salustowicz 2D (S2)} \cite{Vladislavleva2009}
(called Vladislavleva-3 in \cite{McDermott2012}) has similar features as S1, but in two dimensions.

\runintitle{Unwrapped Ball 5D (UB)} \cite{Vladislavleva2009}
is specific by the presence of a fraction and consists of 5 features which all influence the target value.
Again, it does not fit the generalized linear model structure well.

\runintitle{A note on training and testing sampling.}
Originally (i.e. in the referenced articles), some of the benchmarks had different sampling for training and testing data than we present here.
There are two modifications we have made:
\begin{itemize}
    \item For Koza-1, originally there is no testing set, i.e. the same points are used both for training and testing.
    In order to make the results more descriptive, we decided to sample an independent testing set using the same procedure but producing more points (100).
    \item For S1, originally the training sampling is $E[0.05, 10, 0.1]$ and testing sampling is $E[-0.5, 10.5, 0.05]$.
    This means that the range of training data is smaller than the one of testing data.
    Because we want to focus on interpolation rather than extrapolation, we used the bigger of the two ranges, i.e. $[-0.5, 10.5]$ both for training and testing.
    The grid spacing we left at the original values: 0.1 for training and 0.05 for testing.
\end{itemize}

\subsection{Real-World Benchmarks}
The summary of the used real-world benchmarks is in Table \ref{tab:datasets-rw}.
We used random 0.7/0.3 split for training/testing dataset.

\begin{table}[ht]
    \centering
    \caption{Summary of the real-world benchmarks.}
    \label{tab:datasets-rw}
    \begin{tabular}{cccc}
        \toprule
        Name & Dim & \# of datapoints & Ref \\
        \midrule
        ENC & 8 & 768 & \cite{Tsanas2012,uci} \\
        ENH & 8 & 768 & \cite{Tsanas2012,uci} \\
        CCS & 8 & 1030 & \cite{Yeh1998,uci} \\
        ASN & 5 & 1503 & \cite{uci} \\
        \bottomrule
    \end{tabular}
\end{table}

\runintitle{Energy Efficiency (ENC, ENH)} \cite{Tsanas2012}
are datasets regarding energy efficiency of cooling (ENC) and heating (ENH) of buildings, acquired from the UCI repository \cite{uci}.
They were already used as benchmarks in \cite{Arnaldo2015}, where the EFS method was introduced.

\runintitle{Concrete Compressive Strength (CCS)} \cite{Yeh1998}
is a dataset representing a highly non-linear function of concrete age and ingredients, acquired from the UCI repository \cite{uci}.

\runintitle{Airfoil Self-Noise (ASN)},
acquired from the UCI repository \cite{uci}, is a dataset regarding the sound pressure levels of airfoils based on measurements from a wind tunnel.

\subsection{Baseline Algorithms}
In order to provide reasonable baselines for the results of the three SR algorithms, we also computed the results for three classical machine learning algorithms.
The implementations of all three ML algorithms were grabbed from the Python machine learning package, scikit-learn \cite{scikit-learn,scikit-learn-0.17.1}.

\runintitle{Linear Regression (LR)} is an ordinary least-squares multiple linear regression, i.e. without any form of regularization. The model is built just from the original input features.

\runintitle{Random Forest (RF)} is an ensemble regression model made of a number of regression trees, each fitted to a slightly perturbed version of the training data.\footnote{
    For details about the implementation and parameters see \url{http://scikit-learn.org/0.17/modules/generated/sklearn.ensemble.RandomForestRegressor.html}
}
Using the grid search, we tuned the following hyperparameters of the method:
\begin{itemize}
    \item number of trees in the forest with possible values 5, 10, 50, 100, 200, and
    \item number of features to consider when looking for the best split with possible values $N$ and $\sqrt{N}$, where $N$ is the number of features of the dataset.
\end{itemize}

The grid search computes crossvalidation score for each grid point with 3-fold crossvalidation and selects the best settings\footnote{
    For details about the implementation and parameters see \url{http://scikit-learn.org/0.17/modules/generated/sklearn.grid_search.GridSearchCV.html}
}.
The grid search is considered to be a part of the training.

\runintitle{Support Vector Machine for Regression (SVR)}\footnote{
    For details about the implementation and parameters see \url{http://scikit-learn.org/0.17/modules/generated/sklearn.svm.SVR.html}
} with RBF kernel, combined with grid search in in the following hyperparameters:
\begin{itemize}
    \item $C$, the penalty parameter of the error term, with possible values $10^{-3}$, $10^{-2}$, $10^{-1}$, $10^0$, $10^1$, $10^2$, $10^3$, and
    \item $\gamma$, the parameter of the RBF kernel, with possible values $0.01/N$, $0.1/N$, $1/N$, $10/N$, $100/N$, $1000/N$, where $N$ is the number of features of the dataset.
\end{itemize}

The grid search works in the same way as in RF.

\subsection{Settings and Usage of the Algorithms}
The goal is to perform a comparison of the chosen methods as ready-to-use tools.
Therefore we didn't modify to the code of the algorithms\footnote{
    The only exception is EFS: we changed the \texttt{round} variable to \texttt{false} (which was originally hard-coded to \texttt{true}) according to the issue we opened on the algorithm's GitHub repository, see \url{https://github.com/flexgp/efs/issues/1}.
}, and we left all of the settings at their default values.
See more details below.

Additionally, because the default function set of GPTIPS is very limited, we added a second version of GPTIPS, which we refer to as mGPTIPS, with the function set as close as possible to that of EFS
without coding new functions, i.e. using only functions already available (either in MATLAB or in the GPTIPS package).
This is possible because GPTIPS is easily configurable via a config file without the need to modify the code (in contrast to the other methods).
Summary of the function sets of all compared methods is in Table \ref{tab:fn-sets}.

\begin{table}[H]
    \centering
    \begin{threeparttable}
        \caption{
            Function sets of individual algorithms.
            Functions prefixed with ``p'' are protected, \textbf{add3} and \textbf{mult3} are ternary addition and ternary multiplication, respectively.
        }
        \label{tab:fn-sets}
        \begin{tabularx}{.9\textwidth}{*5{>{\centering\arraybackslash}X}@{}}
            \toprule
            function & GPTIPS & mGPTIPS & EFS & FFX \\
            \midrule
            {\bfseries add} & \checkmark & \checkmark & \checkmark &
            \checkmark\tnote{a} \\
            {\bfseries add3} & \checkmark & & & \\
            {\bfseries sub} & \checkmark & \checkmark & \checkmark &
            \checkmark\tnote{a} \\
            {\bfseries mult} & \checkmark & \checkmark & \checkmark & \checkmark
            \\
            {\bfseries mult3} & \checkmark & & & \\
            {\bfseries div} & & \checkmark\tnote{p} & \checkmark\tnote{p} & \checkmark\tnote{b} \\
            {\bfseries sqrt} & & \checkmark\tnote{p} & \checkmark\tnote{p} & \checkmark\tnote{c} \\
            {\bfseries square} & & \checkmark & \checkmark & \\
            {\bfseries cube} & & \checkmark & \checkmark & \\
            {\bfseries quart} & & & \checkmark & \\
            {\bfseries log} & & \checkmark\tnote{p} & \checkmark\tnote{p} & \checkmark \\
            {\bfseries sin} & & \checkmark & \checkmark & \\
            {\bfseries cos} & & \checkmark & \checkmark & \\
            {\bfseries abs} & & & & \checkmark \\
            $\mathbf{max(0,x-thr)}$ & & & & \checkmark \\
            $\mathbf{min(0,x-thr)}$ & & & & \checkmark \\
            \bottomrule
        \end{tabularx}
        \begin{tablenotes}
            \footnotesize
            \item[a] Only via top-level linear combination.
            \item[b] Only via rational functions trick and sign of exponent of feature variable.
            \item[c] Only of feature variable.
            \item[p] Protected version.
        \end{tablenotes}
    \end{threeparttable}
\end{table}

\runintitle{Parameter values.}
GPTIPS and mGPTIPS use identical default values of parameters, except the function set.
Among the most interesting parameters:
    population size is 100,
    number of generations is 150,
    tournament size is 10,
    fraction of elites is 0.15,
    max. tree depth is 4,
    max. number of genes is 4,
    and the initialization procedure is Ramped half'n'half.

EFS, except for the timeout, has no user-definable settings.
The number of evolved features is determined automatically from the number of features in the data set.
For details of the parameter settings, see the original paper \cite{Arnaldo2015}.

FFX has no user-definable settings.
But it is worth to note that the possible exponents for a variable are -1, -0.5, 0.5 and 1; it is thus impossible for the algorithm to create e.g. a quartic term.

For EFS and FFX, which use regularized linear regression, we left the regularization settings at their default values.

\runintitle{Model training and selection.}
From each run of each algorithm, we need to get a single model.
EFS returns just a single model as a result, that best fits the training data.
We decided to use the same strategy also for FFX and GPTIPS.
In case of FFX, which produces as its output a set of nondominated models with respect to performance on the testing dataset and the number of bases, we provided the same data set as both the training and testing data, and selected the best model with respect to MSE.
GPTIPS also returns a population of models, from which we chose the best one.

Choosing the model with minimal training set error might not be considered a good practice because of possible overfitting to the training set.
Yet, we decided to do so because of the following reasons:
\begin{itemize}
    \item In all three methods, overfitting is constrained by setting hard limits on the expressional complexity and/or by putting soft emphasis on simpler models (pathwise regularized learning, parsimony pressure).
    \item Underfitting usually has more sever effects on performance than overfitting.
\end{itemize}

\runintitle{Timeout.}
Both EFS and GPTIPS support a timeout after which the computation is terminated.
We set it to 10 minutes for both methods.
However, as will be seen in Table~\ref{tab:runtimes}, all runs of all algorithms (including FFX which has no support for timeout) finished before this timeout.

\subsection{Testing Environment}
We used GPTIPS version 2 retrieved from \cite{gptips-source}, FFX in version 1.3.4 retrieved from \cite{ffx-source-1.3.4}.
EFS was retrieved from \cite{efs-source-6d991fa}.

All computations were performed on the same PC with Intel Core 2 Duo E6550 at 2.33 GHz, running 64-bit Ubuntu 15.04.
The environments for the three algorithms were:
    MATLAB version R2014a (8.3.0.532) 64-bit for GPTIPS,
    Java version 1.8.0\_60-b27 for EFS,
    Python version 2.7.9 (built with GCC 4.9.2) for FFX and
    Python version 3.4.3 (built with GCC 4.9.2) for the baseline algorithms.

\subsection{Testing Methodology}
Each artificial dataset with uniform random sampling (i.e. the $U$-type sampling) was independently sampled 100 times.
Artificial datasets with deterministic sampling (i.e. the $E$-type sampling) are used only in the single instance.
Each real-world dataset was randomly and independently split 100 times into training and testing sets using 70~\% and 30~\% of the datapoints respectively.

Each algorithm was run once on each of the dataset instances producing a single model.
The accuracy and complexity of the resulting models are then aggregated and statistically compared.
The only exception is the FFX algorithm on S1 and S2 datasets:
these datasets are sampled deterministically (so there is only one instance for both these datasets) and the FFX algorithm is also deterministic, hence a single run is sufficient for these cases.

\section{Results} \label{sec:results}
In the following subsections, we discuss the results per dataset, some global trends we recognize in the results, the time demands of the methods, and the differences among SR and ML models.

We define the number of nodes as the sum of the numbers of nodes across all basis functions of the model.
We count only the expression trees themselves, i.e. we do not count the additional coefficients and operators related to the top-level linear combination produced by the linear regresssion approach used in the tested algorithms.
These coefficients and operators are not counted because they are fully dependent on the bases themselves (their number) and counting them brings no interesting information.%
\footnote{The number of nodes is used as a simple common measure of complexity accross all the algorithms only for reporting purposes. The individual algorithms use their own measures of complexity to find the best model.}
FFX's \emph{hinge functions}, having a form of $\max(0, x - thr)$ or similar, count as 5 nodes.

Differences between individual methods in terms of the testing RMSE and the model complexity were statistically evaluated using one-sided Mann-Whitney U-test (MWUT) for each pair of algorithms with the Bonferroni correction with the significance level $\alpha = 0.05$.\footnote{
    Nevertheless, the results are robust with respect to $\alpha$: the same significance of the differences were obtained for $alpha$ ranging from 0.001 to 0.1.
}

\subsection{Error and Complexity By Dataset}
In this subsection we discuss the results from the point of view of the achieved RMSE and model complexity in terms of the number of nodes.
Table \ref{tab:rmses} presents median RMSE for individual algorithms (SR and ML) and problems.
The ranks of the algorithms w.r.t. the testing RMSE and the results of MWUT for errors are presented in Table \ref{tab:rmses-stat}.
Table \ref{tab:nodes} presents median complexities (numbers of nodes) for individual algorithms and problems.
The ranks of the algorithms w.r.t. the model complexity and the results of MWUT for model complexities are presented in Table \ref{tab:nodes-stat}.
The model complexities are compared among the SR models only, since the ``number of nodes'' measure does not make sense for ML models.

\begin{table}[ht]
    \centering
    \caption{
        Median RMSEs on testing data.
        The best value in each row is highlighted.
    }
    \label{tab:rmses}
    \begin{tabular}{cccccccc}
        \toprule
                 & GPTIPS & mGPTIPS & EFS & FFX & LR & RF & SVR \\
        \midrule
        Koza-1   & {\bfseries 0.0000} & {\bfseries 0.0000} & 0.1280 & 0.0633 & 0.6140 & 0.2083 & 0.1044 \\
        Korns-11 & 7.8112 & {\bfseries 7.7492} & 7.7922 & 7.7962 & 7.7979 & 7.9049 & 7.7974 \\
        S1       & 0.2908 & 0.1114 & 0.2687 & 0.2941 & 0.3022 & {\bfseries 0.0148} & 0.0600 \\
        S2       & 0.9938 & 1.1537 & 1.1070 & 1.0071 & 1.0066 & {\bfseries 0.2276} & 0.7380 \\
        UB       & 0.1413 & 0.1142 & 0.0757 & 0.0833 & 0.1882 & 0.0692 & {\bfseries 0.0570} \\
        \midrule
        ENC      & 2.9073 & 2.2775 & 1.6398 & 1.7906 & 3.2516 & 1.6329 & {\bfseries 1.2779} \\
        ENH      & 2.5375 & 1.7167 & 0.5455 & 1.0455 & 2.9256 & {\bfseries 0.5099} & 0.6737 \\
        CCS      & 8.7618 & 7.1780 & 6.4293 & 5.9860 & 10.523 & {\bfseries 5.1694} & 10.026 \\
        ASN      & 4.1384 & 4.0034 & 3.6232 & 3.5804 & 4.8160 & {\bfseries 1.8391} & 6.0543 \\
        \bottomrule
    \end{tabular}
\end{table}

\begin{table}[ht]
    \centering
    \caption{
        Statistical ranking of RMSEs.
        Left columns show the rank of the algorithm. The title of right columns, ``ssbt'', stands for \emph{statistically significantly better than}, and they show algorithms that were statistically significantly worse as judged by the Mann-Whitney U-test.
        The significance level after the Bonferroni correction for 21 pairs is $\alpha \approxeq 0.0024$.
        The individual algorithms are denoted by their first letter: \textbf{G} for GPTIPS, \textbf{m} for mGPTIPS, \textbf{E} for EFS, \textbf{F} for FFX, \textbf{L} for LR, \textbf{R} for RF, and \textbf{S} for SVR.
    }
    \label{tab:rmses-stat}
    \begin{tabular}{c@{\hspace{2mm}}c@{}c@{\hspace{.7mm}}|@{\hspace{.7mm}}c@{}c@{\hspace{.7mm}}|@{\hspace{.7mm}}c@{}c@{\hspace{.7mm}}|@{\hspace{.7mm}}c@{}c@{\hspace{.7mm}}|@{\hspace{.7mm}}c@{\hspace{1mm}}c@{\hspace{.7mm}}|@{\hspace{.7mm}}c@{}c@{\hspace{.7mm}}|@{\hspace{.7mm}}c@{}c}
        \toprule
        & \multicolumn{2}{c@{\hspace{1.7mm}}|@{\hspace{.7mm}}}{{\bfseries G}PTIPS} & \multicolumn{2}{c@{\hspace{1.7mm}}|@{\hspace{.7mm}}}{{\bfseries m}GPTIPS} & \multicolumn{2}{c@{\hspace{1.7mm}}|@{\hspace{.7mm}}}{{\bfseries E}FS} & \multicolumn{2}{c@{\hspace{1.7mm}}|@{\hspace{.7mm}}}{{\bfseries F}FX} & \multicolumn{2}{c@{\hspace{1.7mm}}|@{\hspace{.7mm}}}{{\bfseries L}R} & \multicolumn{2}{c@{\hspace{1.7mm}}|@{\hspace{.7mm}}}{{\bfseries R}F} & \multicolumn{2}{c}{{\bfseries S}VR} \\
        & {\scriptsize rank} & {\scriptsize ssbt} & {\scriptsize rank} & {\scriptsize ssbt} & {\scriptsize rank} & {\scriptsize ssbt} & {\scriptsize rank} & {\scriptsize ssbt} & {\scriptsize rank} & {\scriptsize ssbt} & {\scriptsize rank} & {\scriptsize ssbt} & {\scriptsize rank} & {\scriptsize ssbt} \\
        \midrule
        Koza-1   & 1-2   & EFLRS & 1-2  & EFLRS  & 5-6   & L     & 3     & LRS   & 7   &    & 5-6 & L      & 4   & LR \\
        Korns-11 & 2-6   & R     & 1    & GEFLRS & 2-6   & R     & 2-6   & R     & 2-6 & R  & 7   &        & 2-6 & R \\
        S1       & 5     & FL    & 3    & GEFL   & 4     & GFL   & 6     & L     & 7   &    & 1   & GmEFLS & 2   & GmEFL \\
        S2       & 3     & EFL   & 6-7  &        & 6-7   &       & 5     & E     & 4   & EF & 1   & GmEFLS & 2   & GmEFL \\
        UB       & 6     & L     & 5    & GL     & 3     & GmFL  & 4     & GmL   & 7   &    & 2   & GmEFL  & 1   & GmEFLR \\
        \midrule
        ENC      & 6     & L     & 5    & GL     & 2-3   & GmFL  & 4     & GmL   & 7   &    & 2-3 & GmFL   & 1   & GmEFLR \\
        ENH      & 6     & L     & 5    & GL     & 1-2   & GmFLS & 4     & GmL   & 7   &    & 1-2 & GmFLS  & 3   & GmFL \\
        CCS      & 5     & LS    & 4    & GLS    & 3     & GmLS  & 2     & GmELS & 7   &    & 1   & GmEFLS & 6   & L \\
        ASN      & 5     & LS    & 4    & GLS    & 2-3   & GmLS  & 2-3   & GmLS  & 6   & S  & 1   & GmEFLS & 7   & \\
        \bottomrule
    \end{tabular}
\end{table}

\begin{table}[ht]
    \centering
    \caption{Median number of nodes for each algorithm and dataset.}
    \label{tab:nodes}
    \begin{tabular}{ccccc}
        \toprule
        & GPTIPS & mGPTIPS & EFS & FFX \\
        \midrule
        Koza-1   & 33 & 14 & {\bfseries 11} & 35 \\
        Korns-11 & 63 & 17 & 69 & {\bfseries 14} \\
        S1       & 52 & 23 & 12 & {\bfseries 10} \\
        S2       & 53 & 25 & 28 & {\bfseries 1} \\
        UB       & 36.5 & {\bfseries 10.5} & 66 & 105 \\
        \midrule
        ENC      & 48 & {\bfseries 25} & 108 & 136 \\
        ENH      & 47.5 & {\bfseries 26} & 105 & 146 \\
        CCS      & 43 & {\bfseries 23} & 108 & 474.5 \\
        ASN      & 58 & {\bfseries 30} & 67 & 52.5 \\
        \bottomrule
    \end{tabular}
\end{table}

\begin{table}[ht]
    \centering
    \caption{
        Statistical ranking of complexities (number of nodes).
        Left columns show the rank of the algorithm. The title of right columns, ``ssbt'', stands for \emph{statistically significantly better than}, and they show algorithms that were statistically significantly worse as judged by the Mann-Whitney U-test. 
        The significance level after the Bonferroni correction for 6 pairs is $\alpha \approxeq 0.0083$.
        The individual algorithms are denoted by their first letter: \textbf{G} for GPTIPS, \textbf{m} for mGPTIPS, \textbf{E} for EFS, and \textbf{F} for FFX.
    }
    \label{tab:nodes-stat}
    \begin{tabular}{c@{\hspace{2mm}}c@{}c@{\hspace{.7mm}}|@{\hspace{.7mm}}c@{}c@{\hspace{.7mm}}|@{\hspace{.7mm}}c@{}c@{\hspace{.7mm}}|@{\hspace{.7mm}}c@{}c}
        \toprule
        & \multicolumn{2}{c@{\hspace{1.7mm}}|@{\hspace{.7mm}}}{{\bfseries G}PTIPS} & \multicolumn{2}{c@{\hspace{1.7mm}}|@{\hspace{.7mm}}}{{\bfseries m}GPTIPS} & \multicolumn{2}{c@{\hspace{1.7mm}}|@{\hspace{.7mm}}}{{\bfseries E}FS} & \multicolumn{2}{c}{{\bfseries F}FX} \\
        & {\scriptsize rank} & {\scriptsize ssbt} & {\scriptsize rank} & {\scriptsize ssbt} & {\scriptsize rank} & {\scriptsize ssbt} & {\scriptsize rank} & {\scriptsize ssbt} \\
        \midrule
        Koza-1   & 3     &     & 2   & FG  & 1     & FGm & 4     &     \\
        Korns-11 & 3     & E   & 2   & EG  & 4     &     & 1     & EG  \\
        S1       & 4     &     & 3   & G   & 2     & Gm  & 1     & EGm \\
        S2       & 4     &     & 2   & G   & 3     & G   & 1     & EGm \\
        UB       & 2     & EF  & 1   & EFG & 3     & F   & 4     &     \\
        \midrule
        ENC      & 2     & EF  & 1   & EFG & 3     & F   & 4     &     \\
        ENH      & 2     & EF  & 1   & EFG & 3     & F   & 4     &     \\
        CCS      & 2     & EF  & 1   & EFG & 3     & F   & 4     &     \\
        ASN      & 3     & E   & 1   & EFG & 4     &     & 2    & E   \\
        \bottomrule
    \end{tabular}
\end{table}

\begin{figure}[ht]
    \centering
    \includegraphics[scale=.305, trim=0 -18mm 0 0]{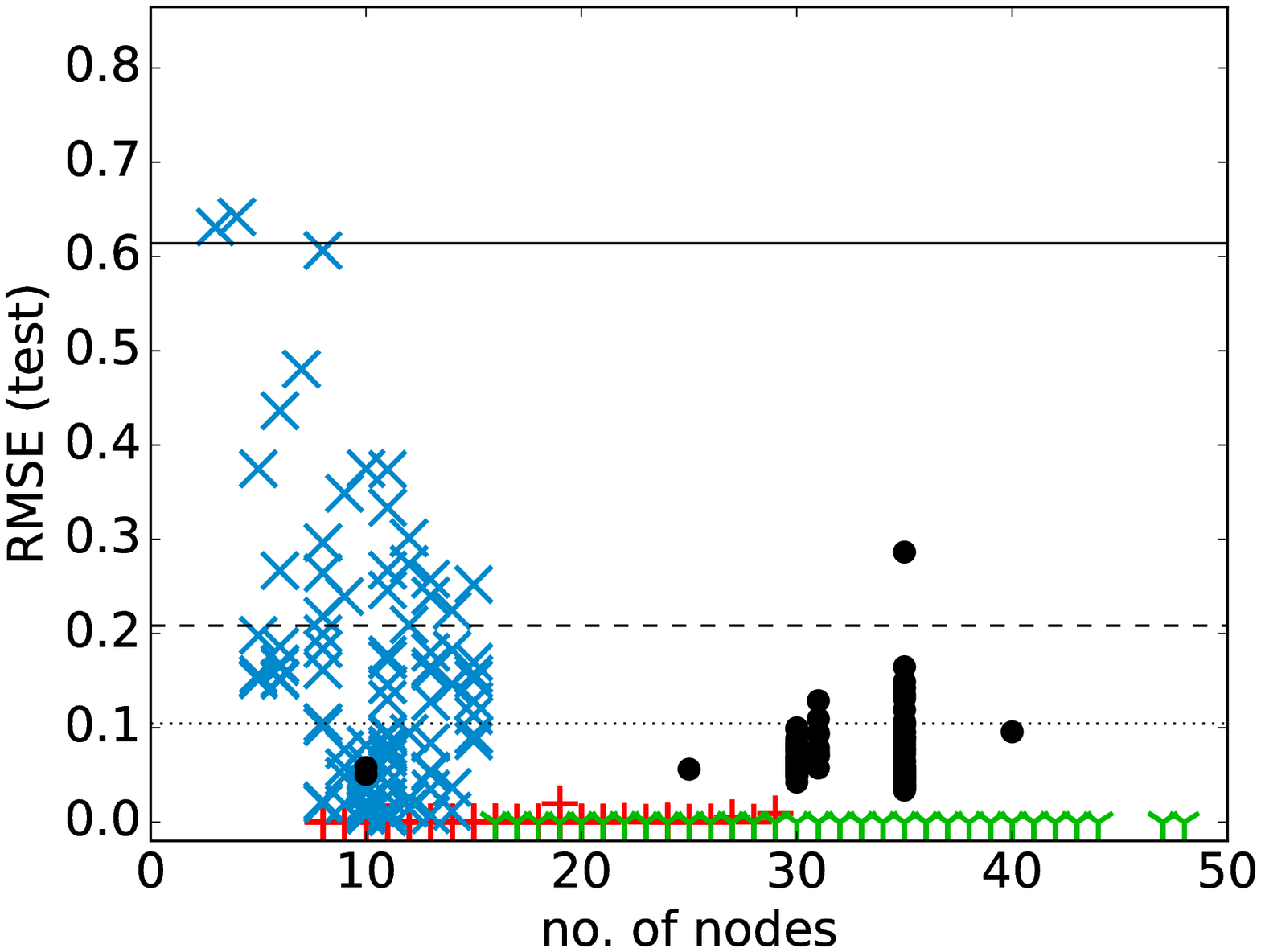}
    \includegraphics[scale=.31, trim=0 0 0 0]{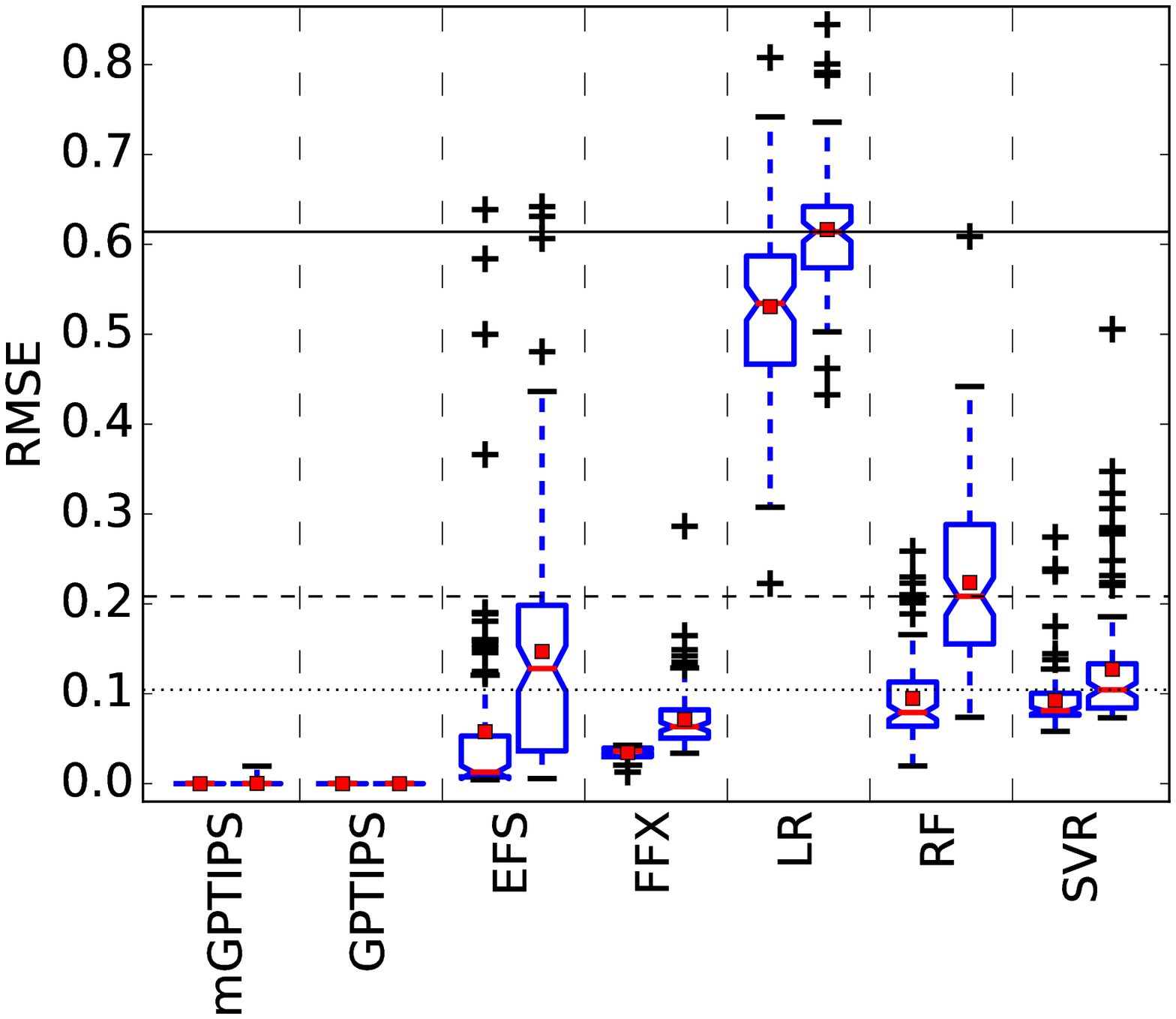}
    \caption{
        Complexity-performance plots (left) and box plots of training and testing errors (right) for the Koza-1 dataset.
        \plotslegend
    }
    \label{fig:koza1}
\end{figure}
\begin{figure}[ht]
    \centering
    \includegraphics[scale=.305, trim=0 -18mm 0 0]{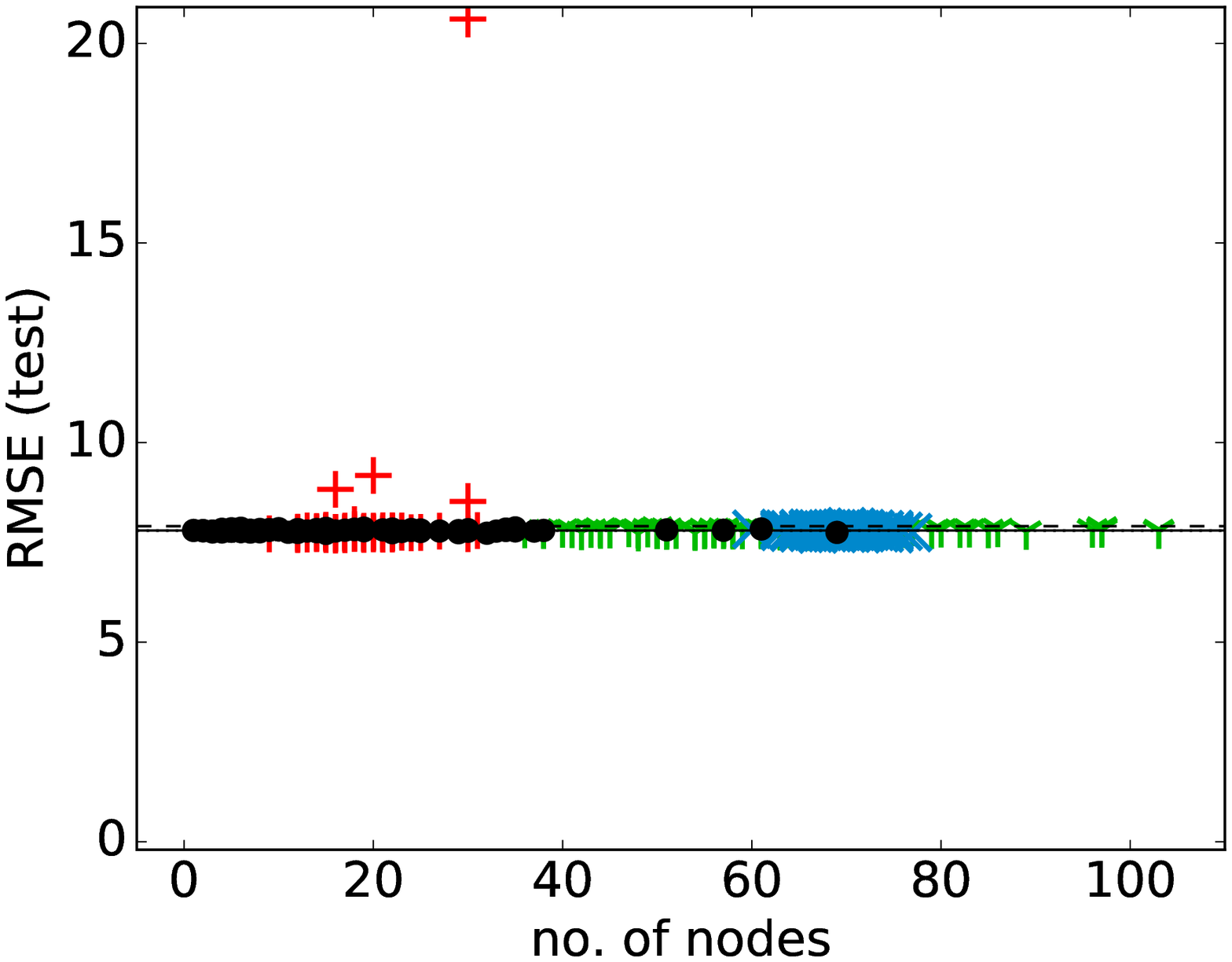}
    \includegraphics[scale=.31, trim=0 0 0 0]{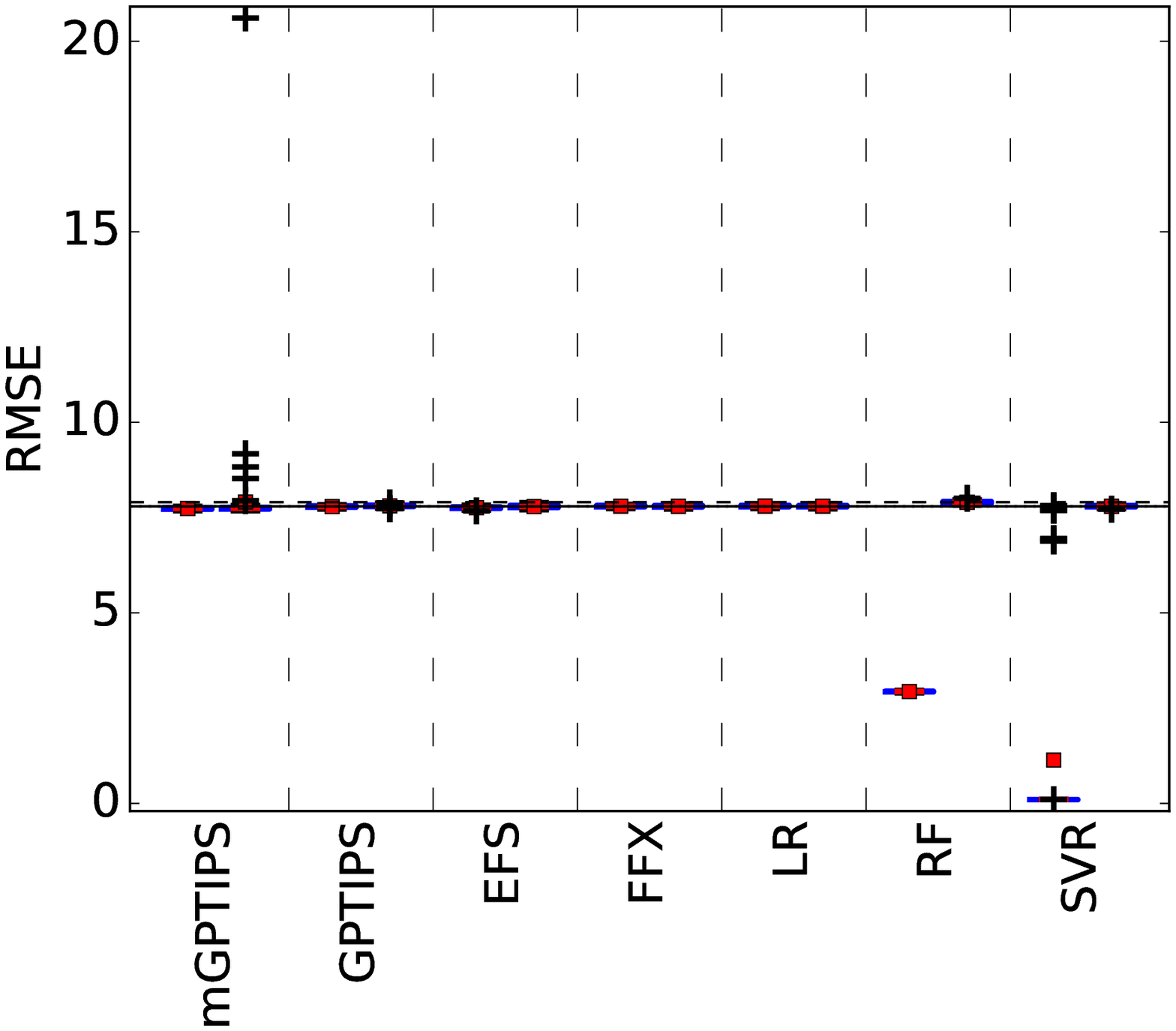}
    \includegraphics[scale=.305, trim=0 -18mm 0 0]{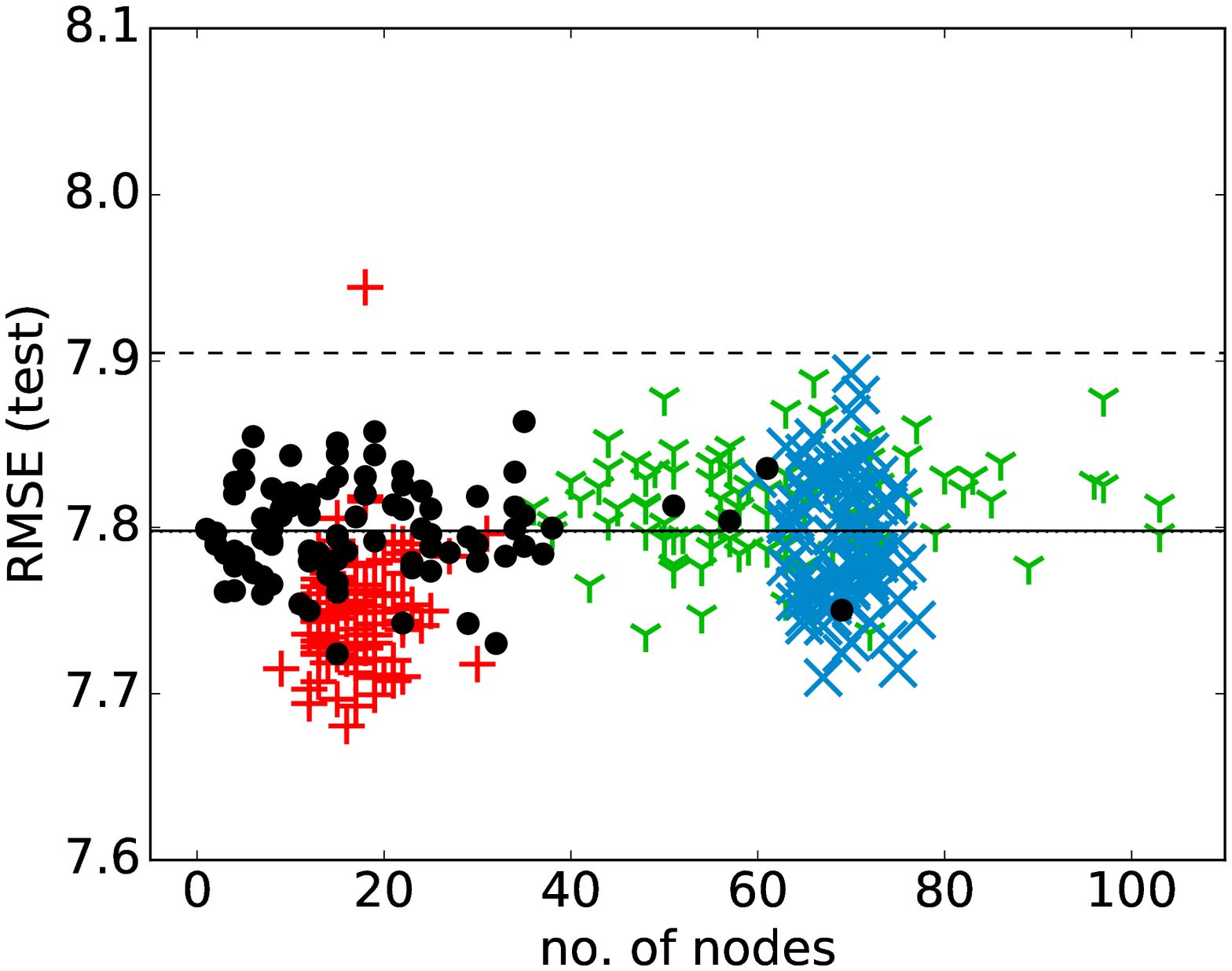}
    \includegraphics[scale=.31, trim=0 0 0 0]{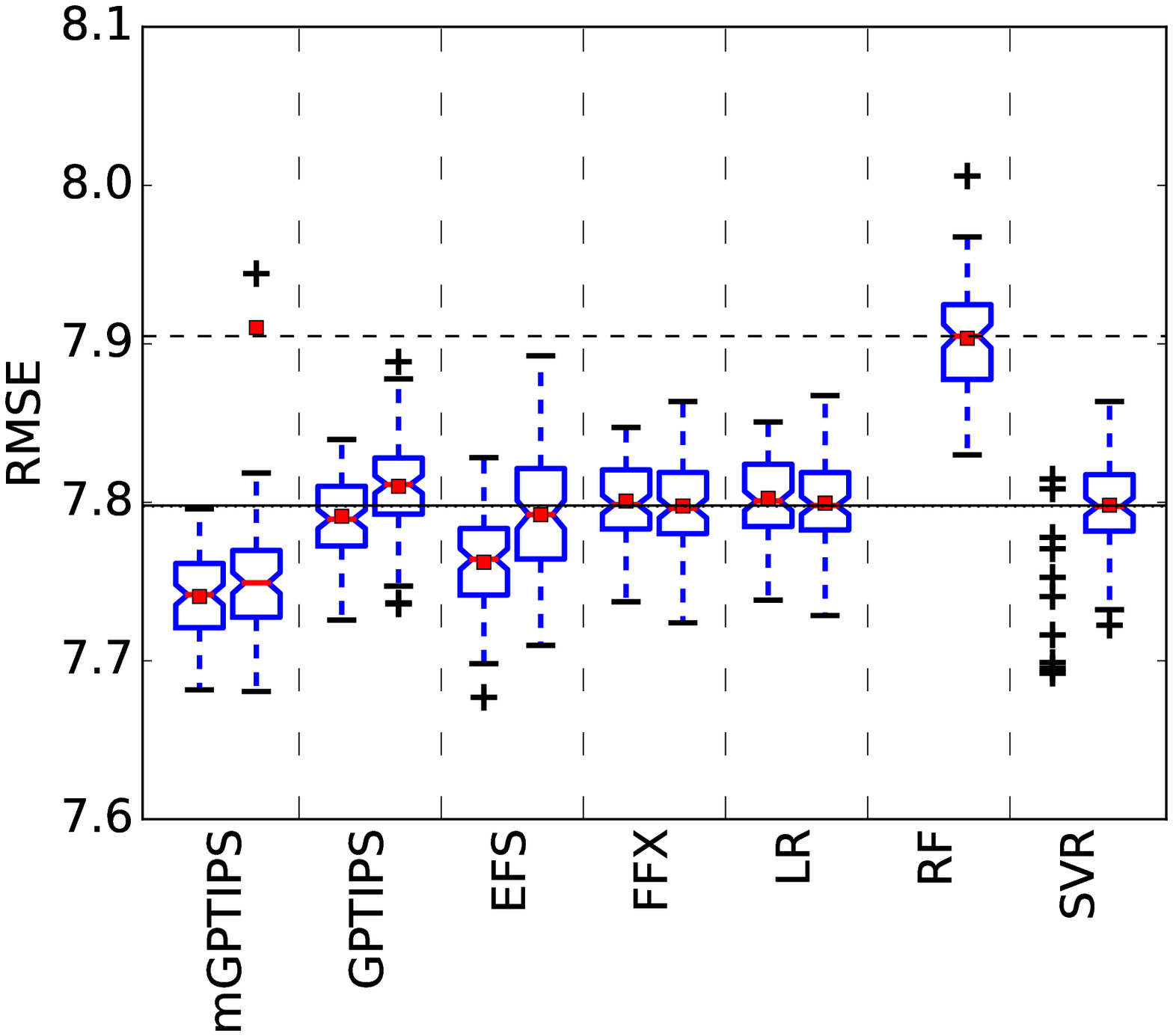}
    \caption{
        Complexity-performance plots (both left) and box plots of training and testing errors (both right) for the Korns-11 dataset.
        The upper plots display the whole results, the lower ones are zoom on the dense area around RMSE = 7.8.
        \plotslegend
    }
    \label{fig:korns11}
\end{figure}
\begin{figure}[ht]
    \centering
    \includegraphics[scale=.305, trim=0 -18mm 0 0]{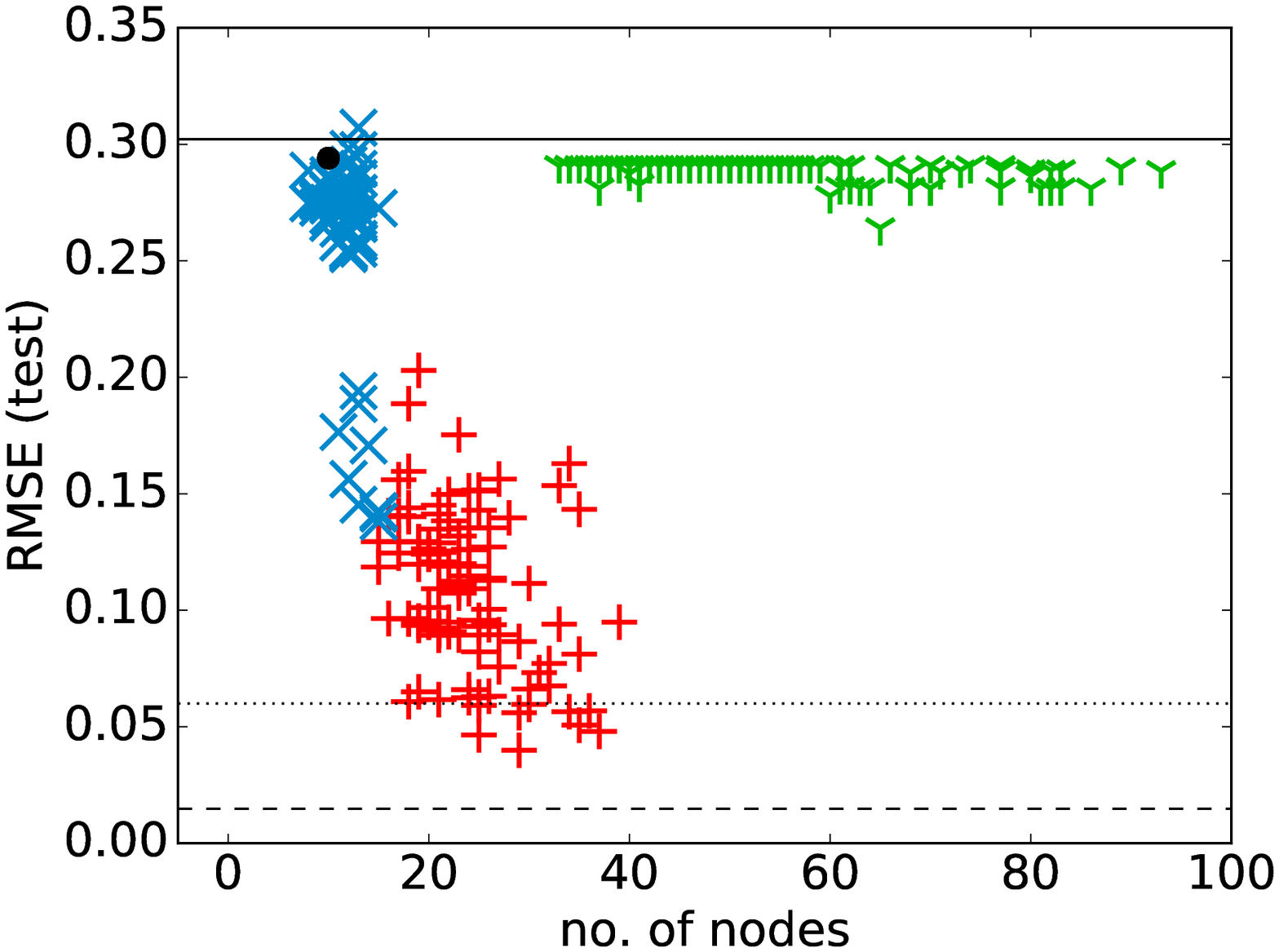}
    \includegraphics[scale=.31, trim=0 0 0 0]{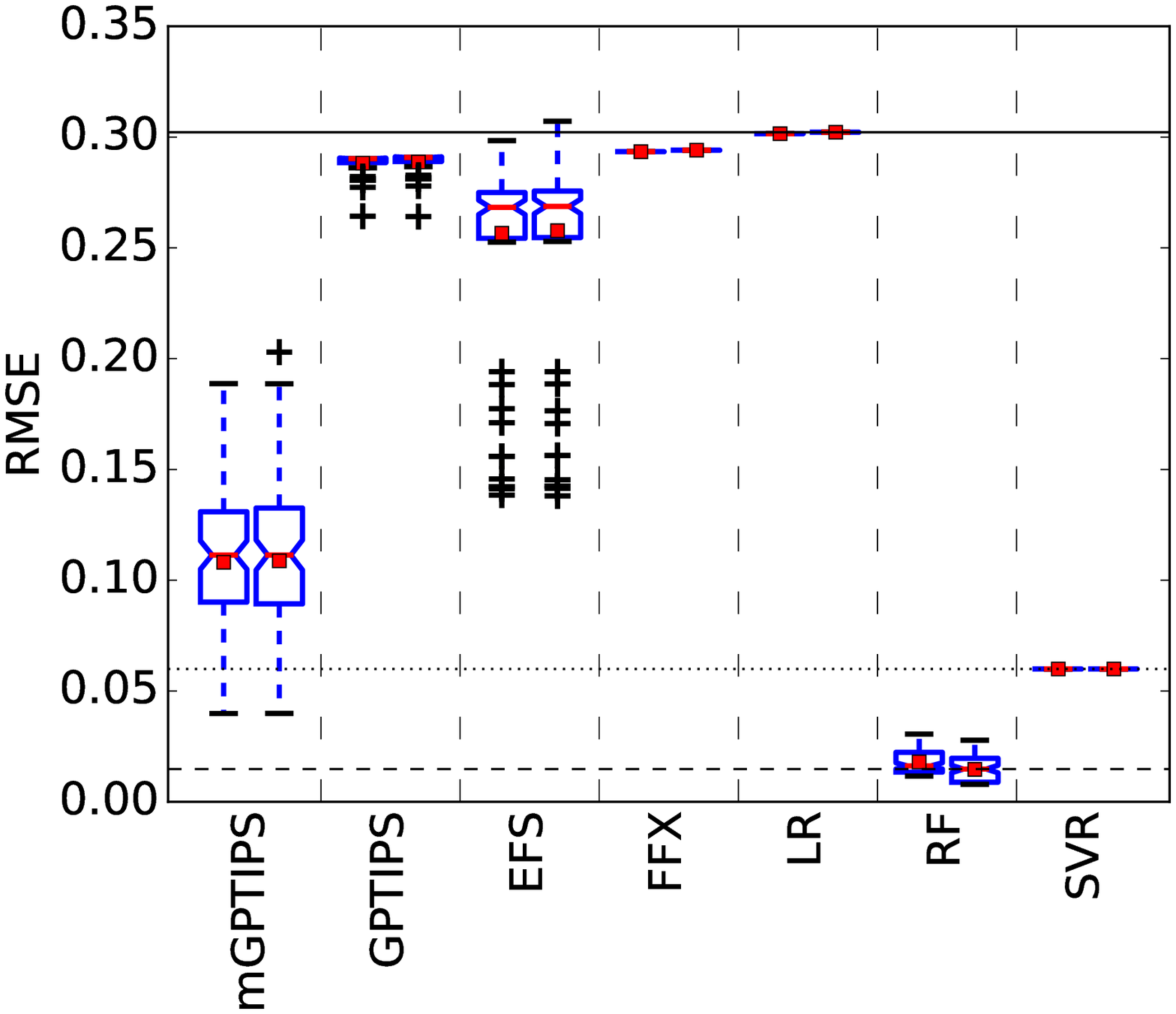}
    \caption{
        Complexity-performance plots (left) and box plots of training and testing errors (right) for the S1 dataset.
        FFX has only a single point because both the sampling of this dataset and FFX are deterministic.
        \plotslegend
    }
    \label{fig:s1}
\end{figure}
\begin{figure}[ht]
    \centering
    \includegraphics[scale=.305, trim=0 -18mm 0 0]{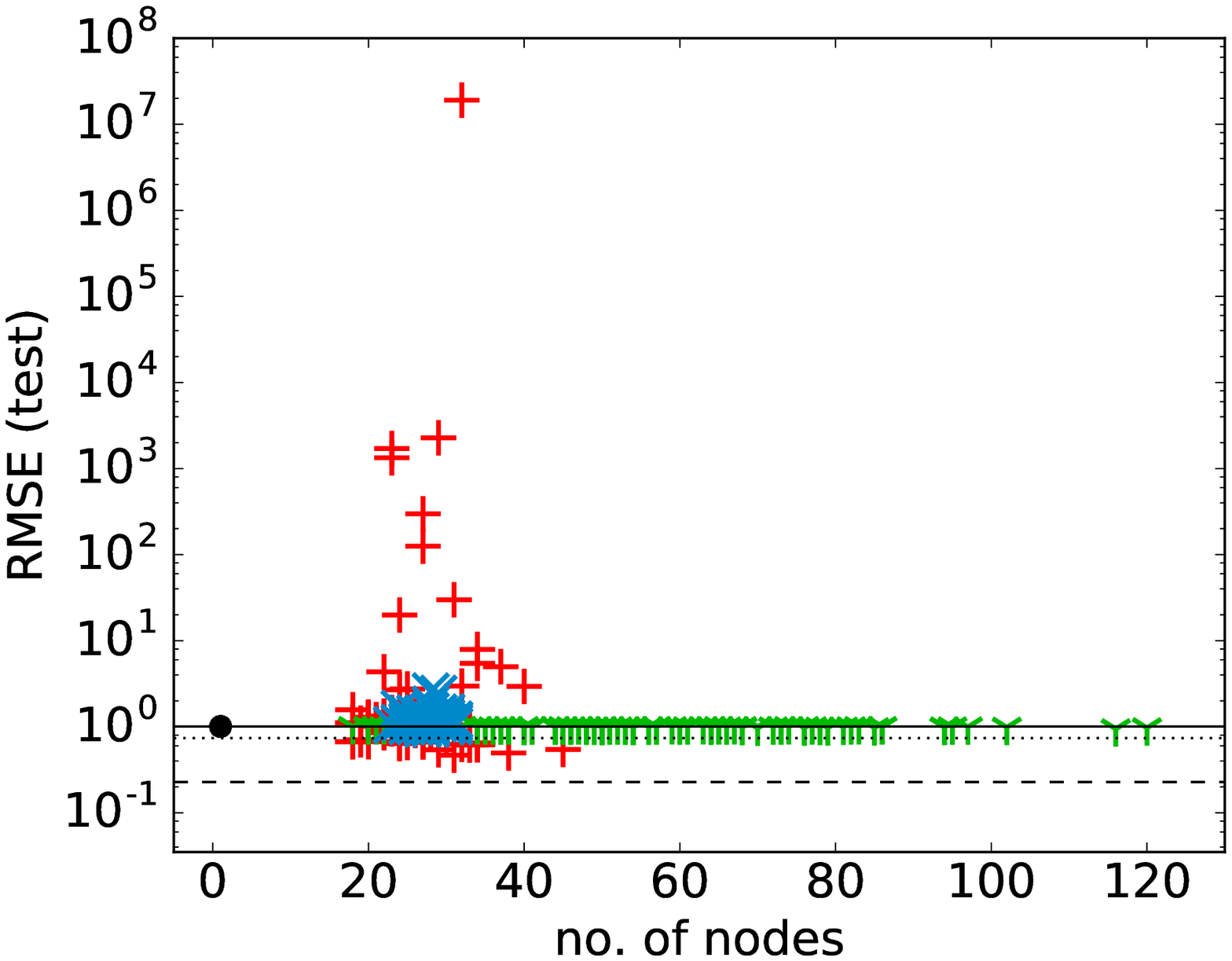}
    \includegraphics[scale=.31, trim=0 0 0 0]{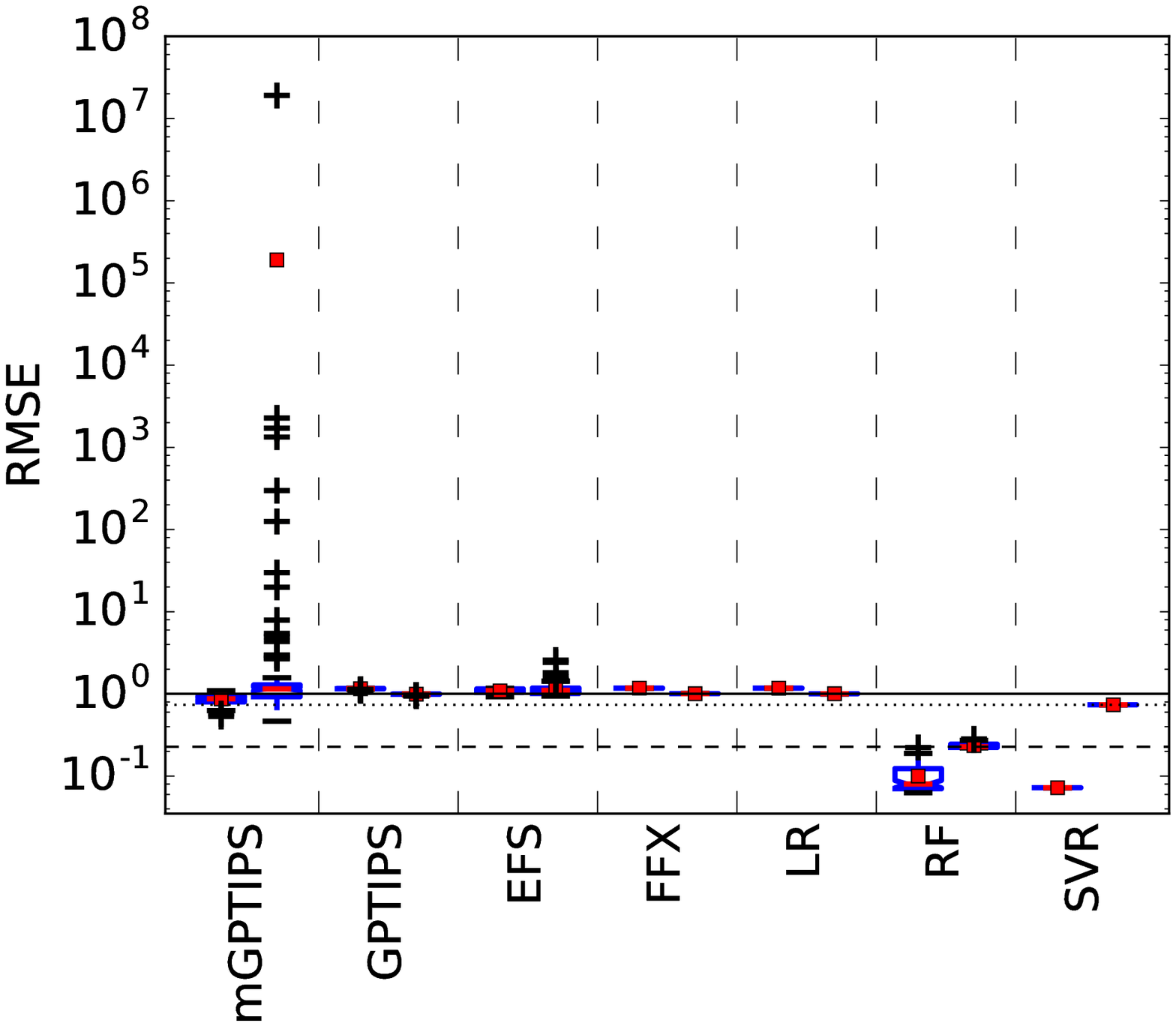}
    \includegraphics[scale=.305, trim=0 -18mm 0 0]{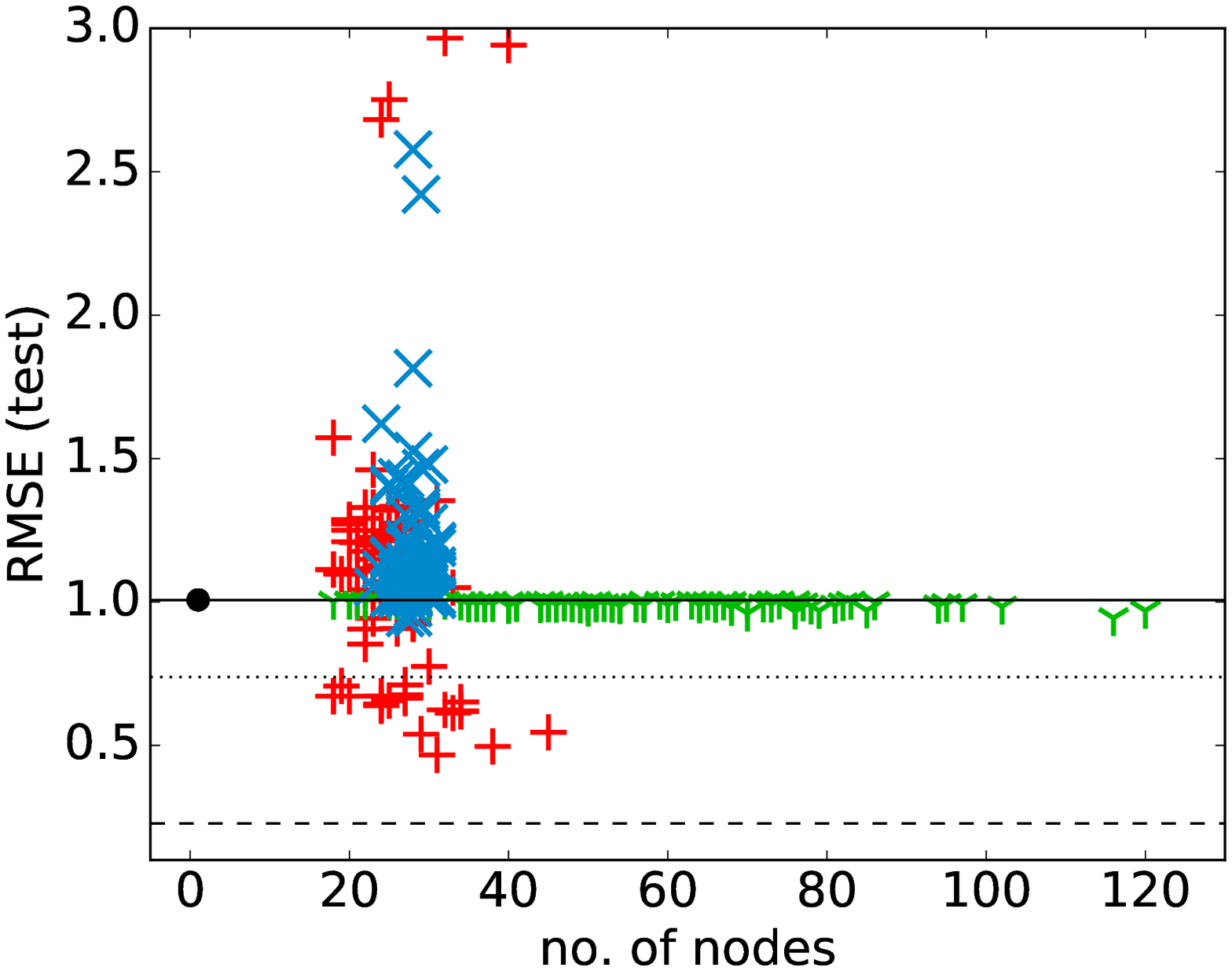}
    \includegraphics[scale=.31, trim=0 0 0 0]{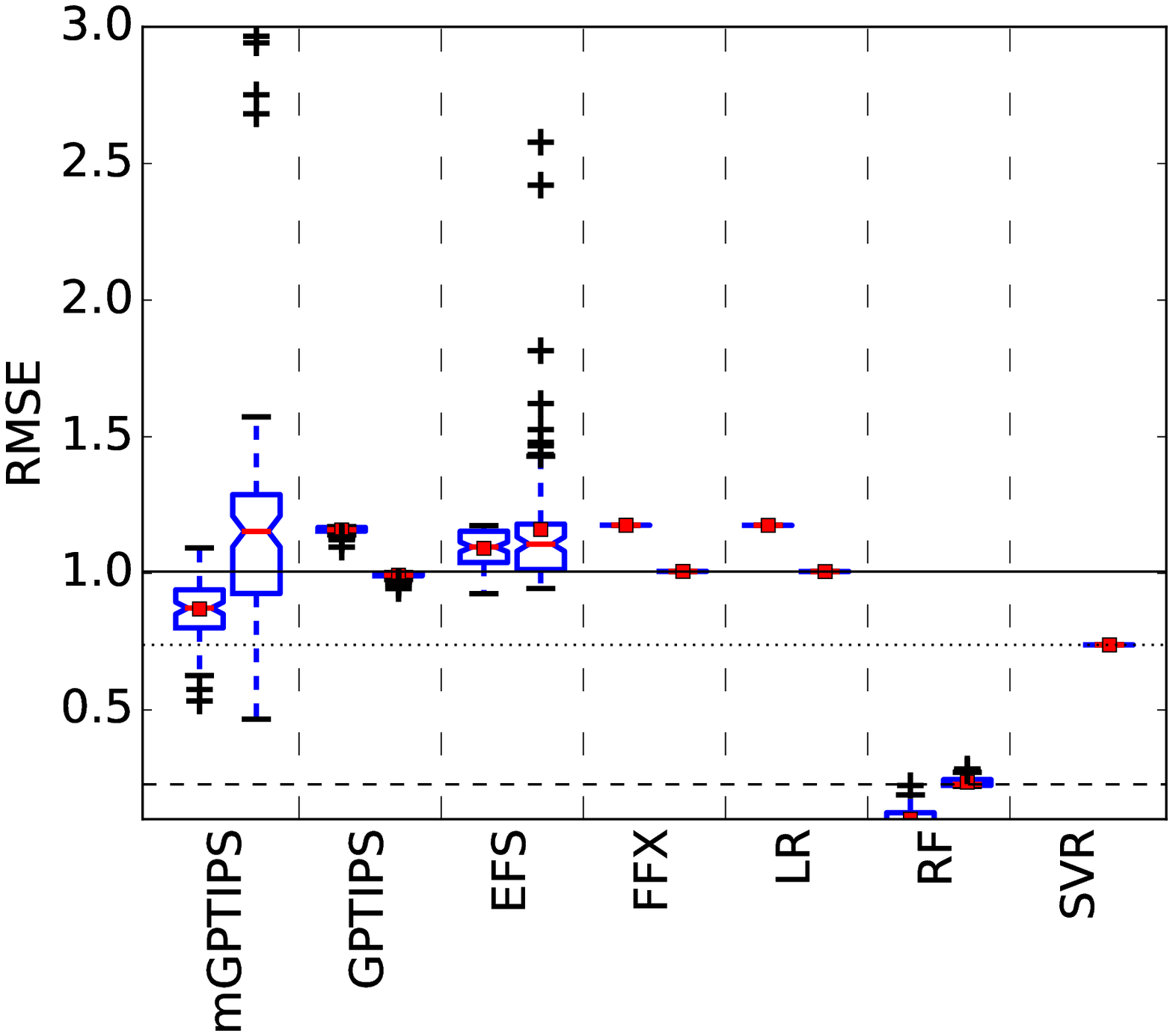}
    \caption{
        Complexity-performance plots (both left) and box plots of training and testing errors (both right) for the S2 dataset.
        FFX has only a single point because both the sampling of this dataset and FFX are deterministic.
        The upper plots display the whole results, the lower ones are zoom on the dense area around RMSE = 1.
        \plotslegend
    }
    \label{fig:s2}
\end{figure}
\begin{figure}[ht]
    \centering
    \includegraphics[scale=.305, trim=0 -18mm 0 0]{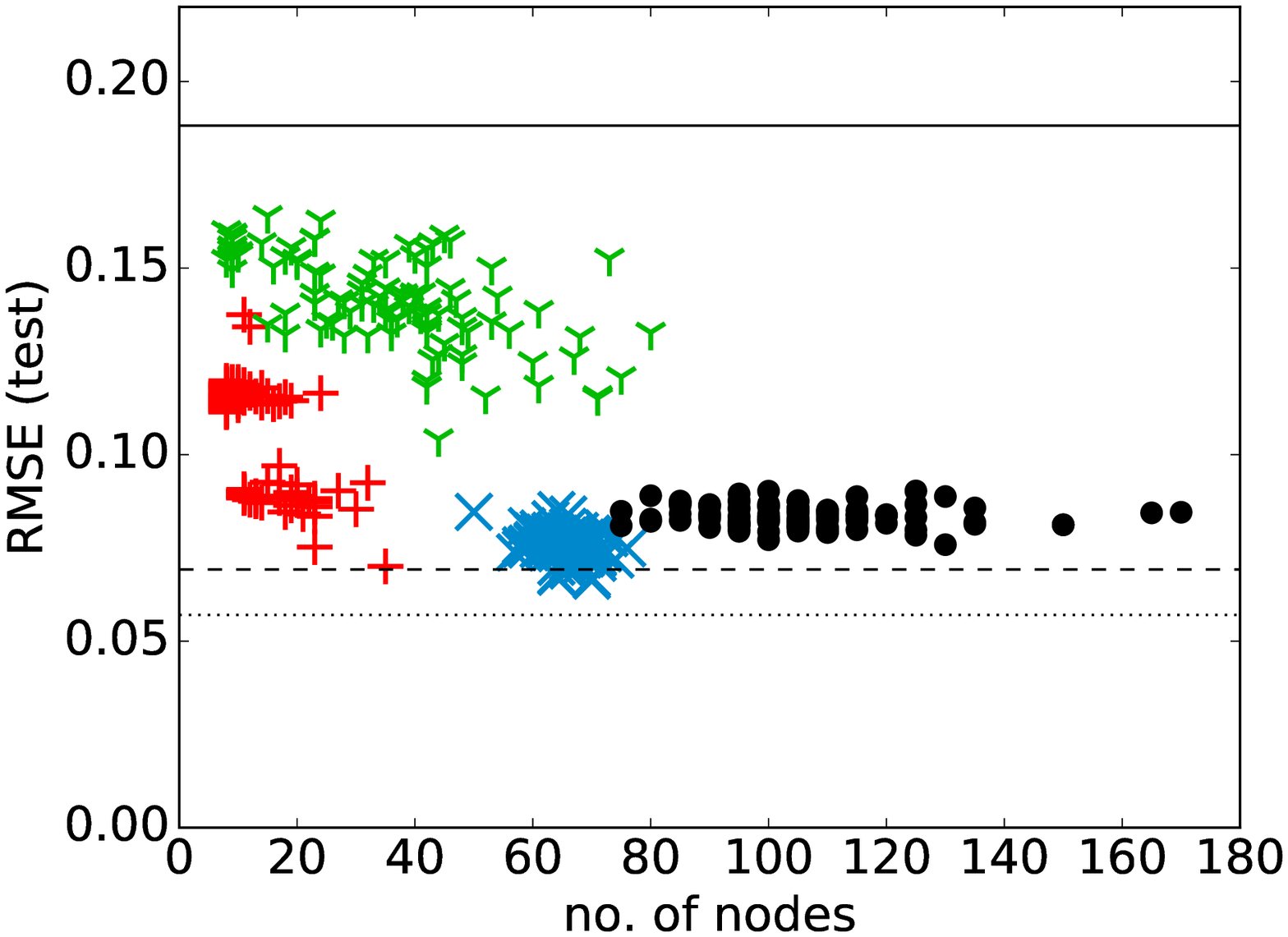}
    \includegraphics[scale=.31, trim=0 0 0 0]{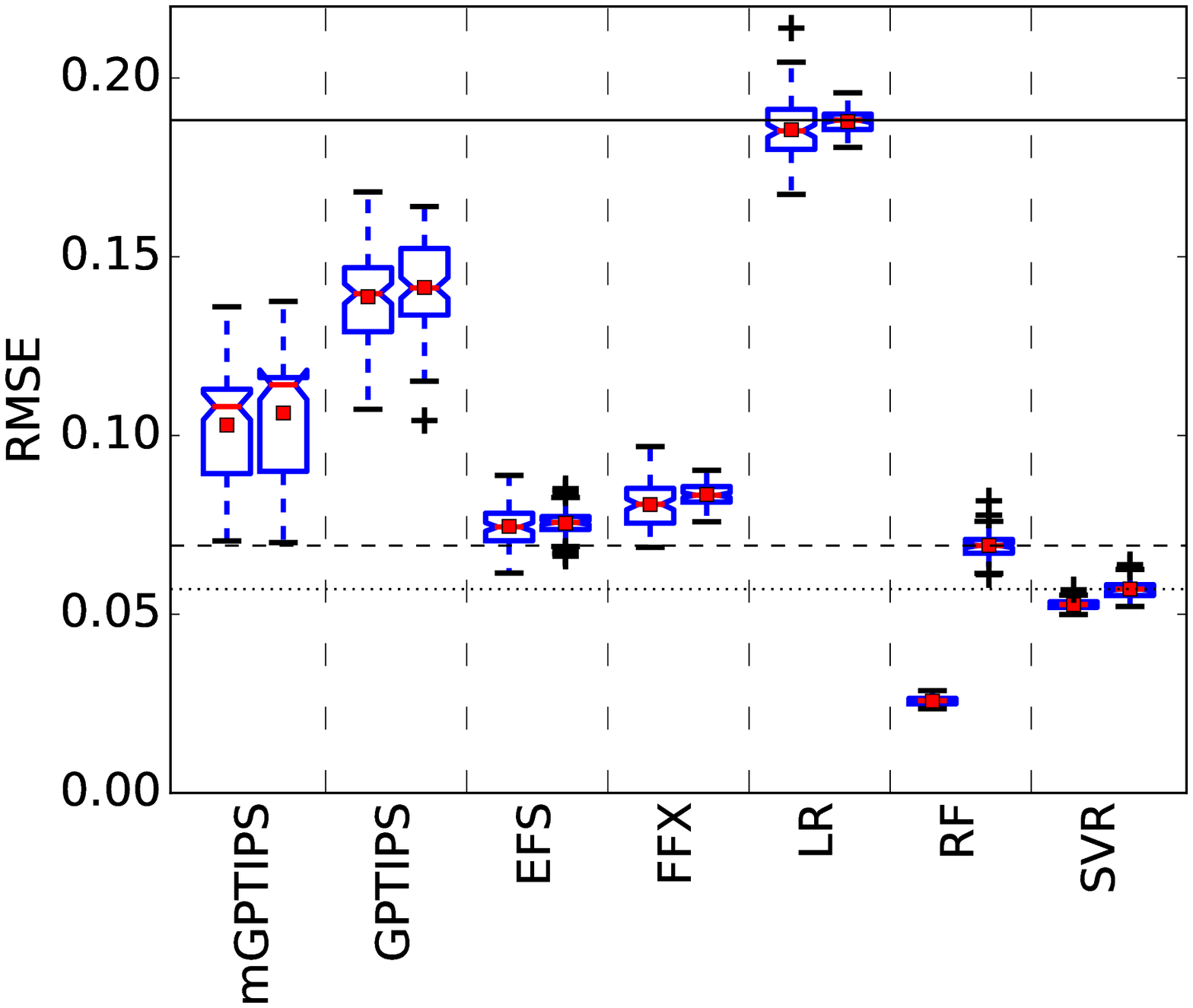}
    \caption{
        Complexity-performance plots (left) and box plots of training and testing errors (right) for the UB dataset.
        \plotslegend
    }
    \label{fig:ub}
\end{figure}

\begin{figure}[ht]
    \centering
    \includegraphics[scale=.305, trim=0 -18mm 0 0]{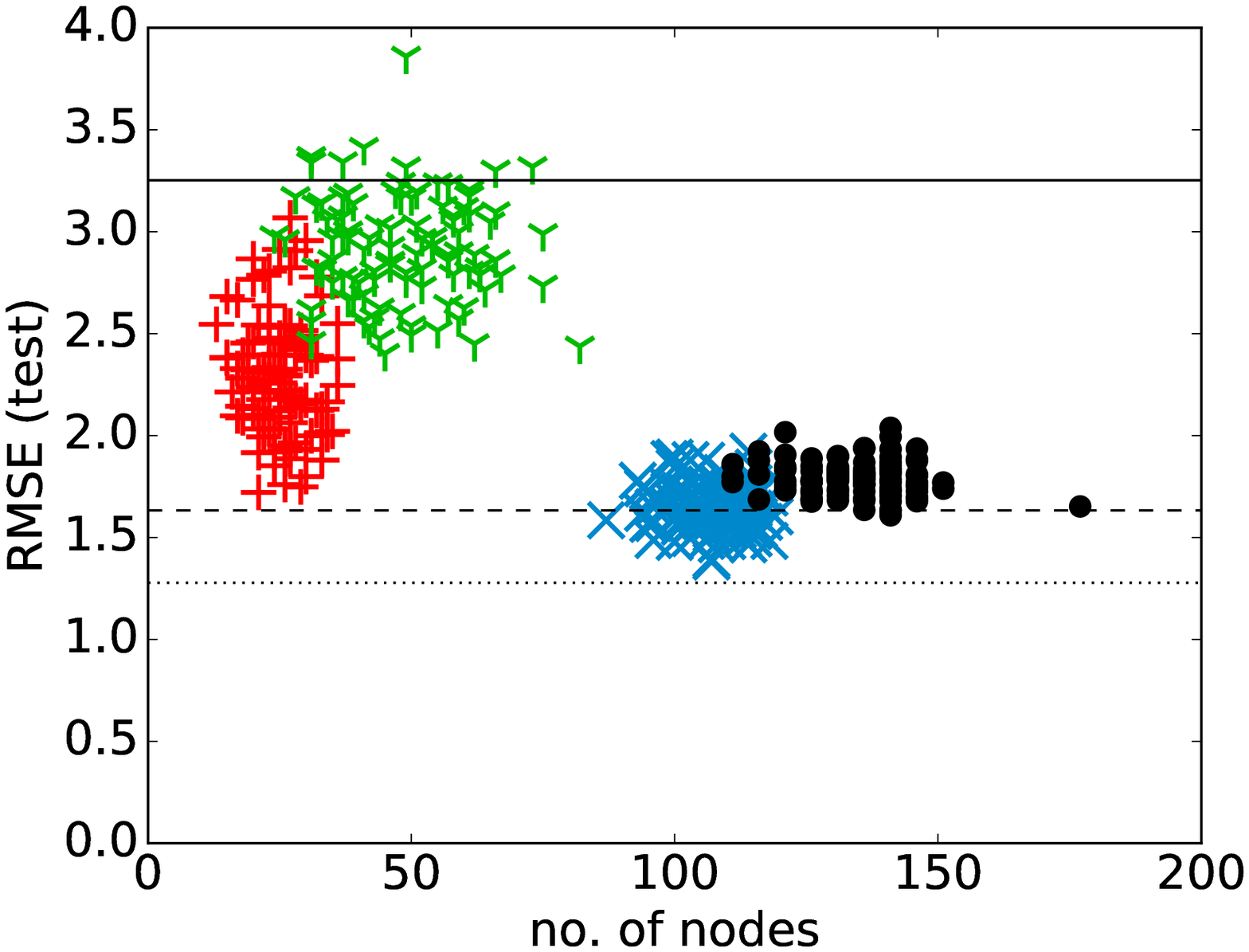}
    \includegraphics[scale=.31, trim=0 0 0 0]{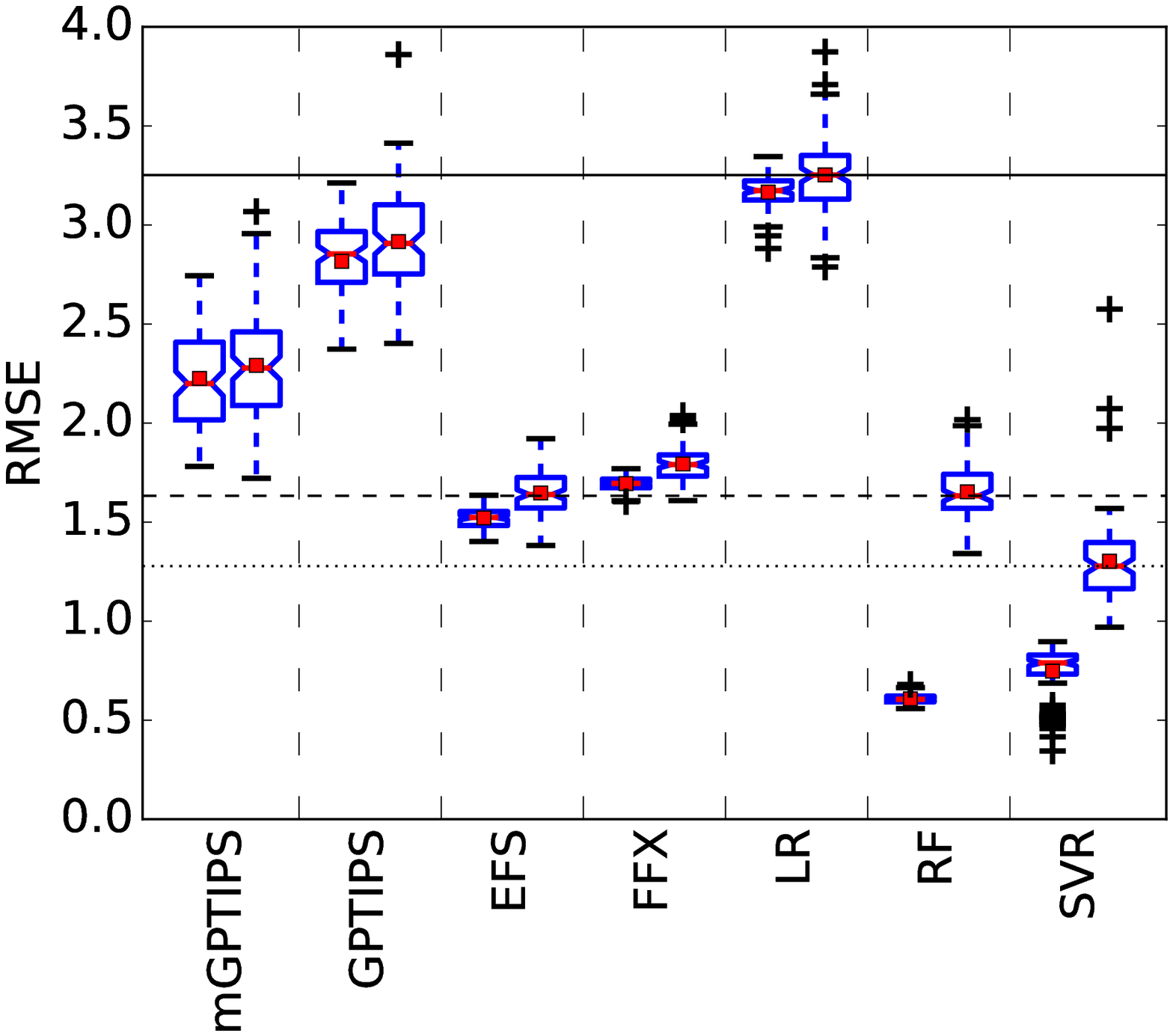}
    \caption{
        Complexity-performance plots (left) and box plots of training and testing errors (right) for the ENC dataset.
        \plotslegend
    }
    \label{fig:enc}
\end{figure}
\begin{figure}[ht]
    \centering
    \includegraphics[scale=.305, trim=0 -18mm 0 0]{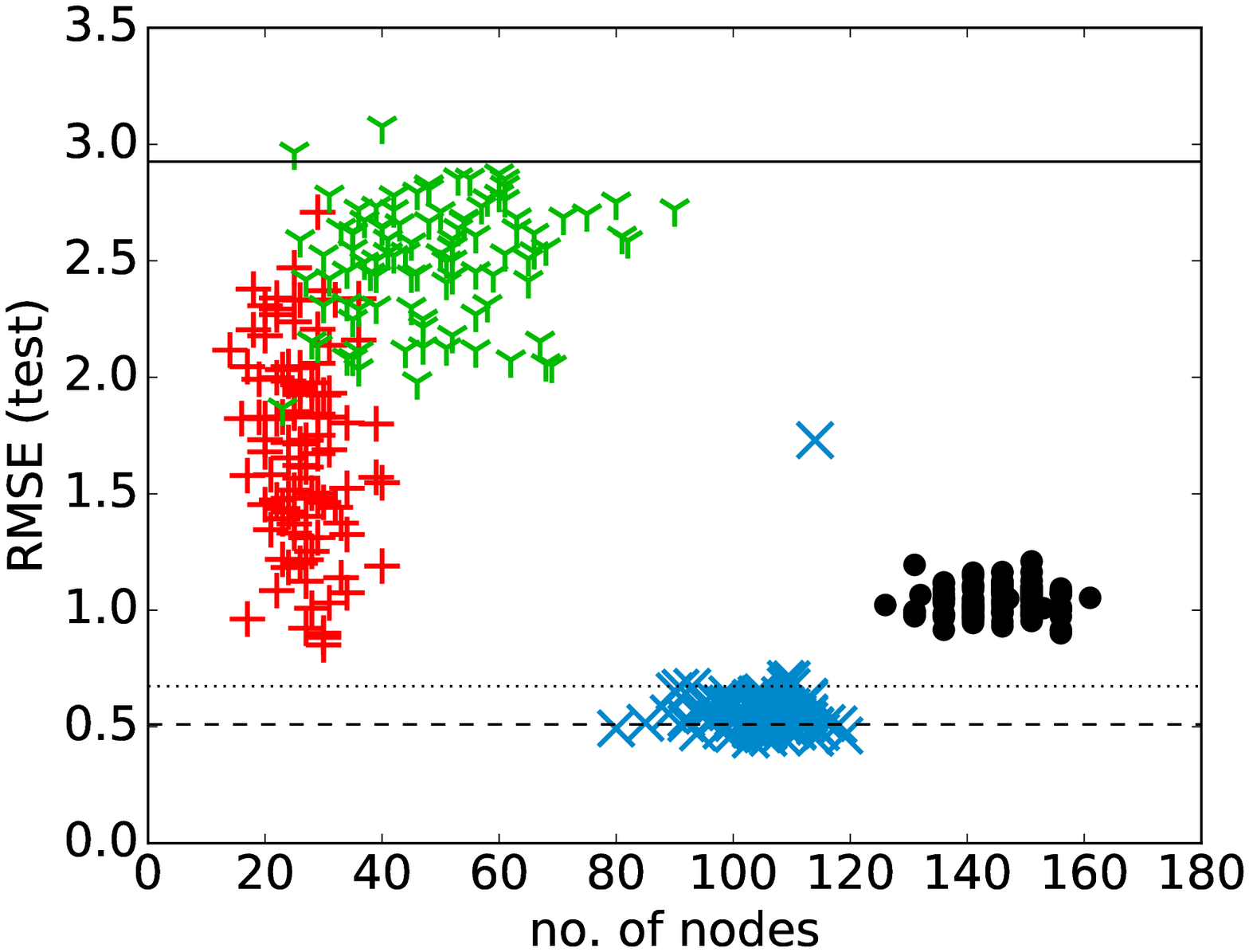}
    \includegraphics[scale=.31, trim=0 0 0 0]{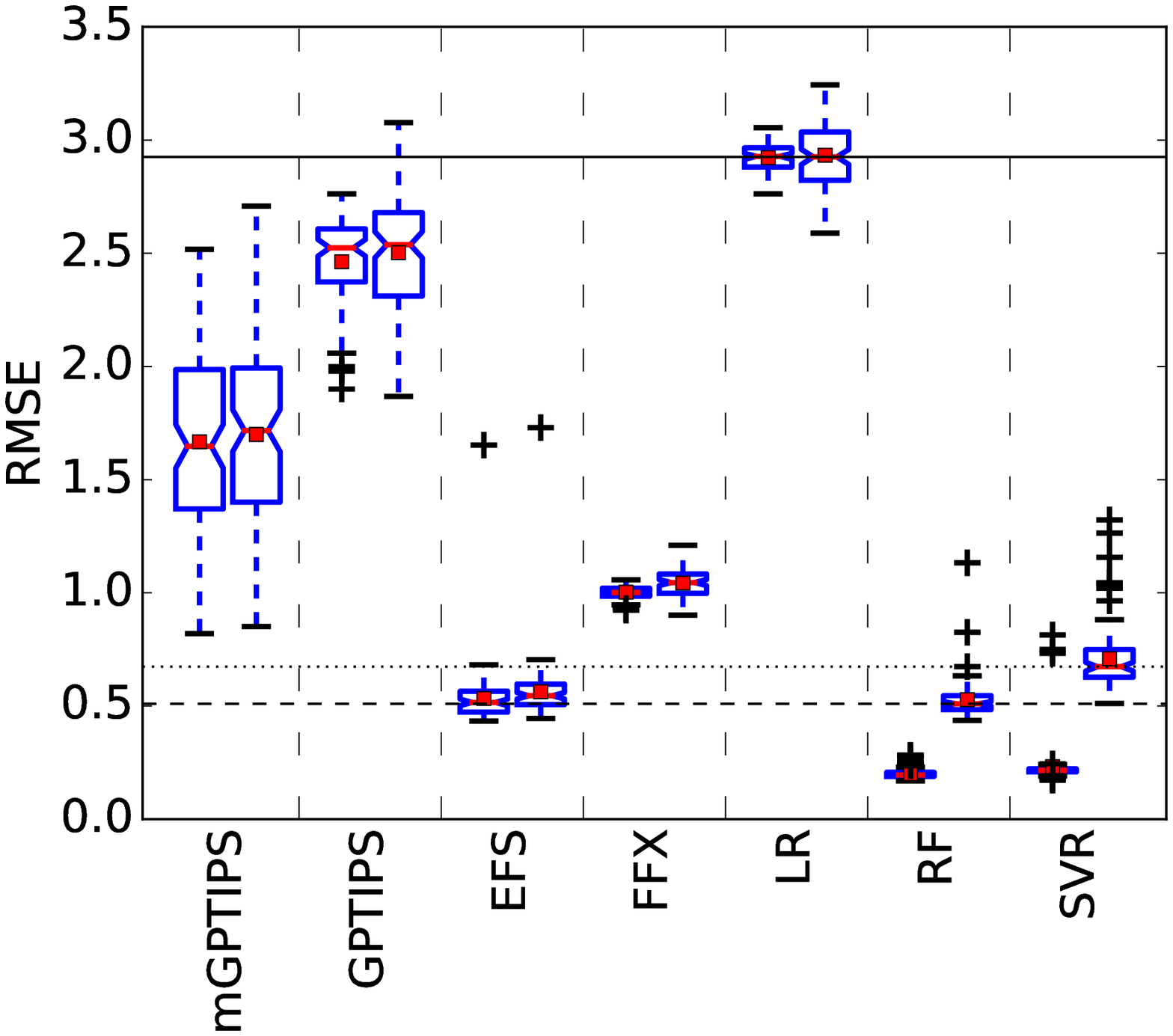}
    \caption{
        Complexity-performance plots (left) and box plots of training and testing errors (right) for the ENH dataset.
        \plotslegend
    }
    \label{fig:enh}
\end{figure}
\begin{figure}[ht]
    \centering
    \includegraphics[scale=.305, trim=0 -18mm 0 0]{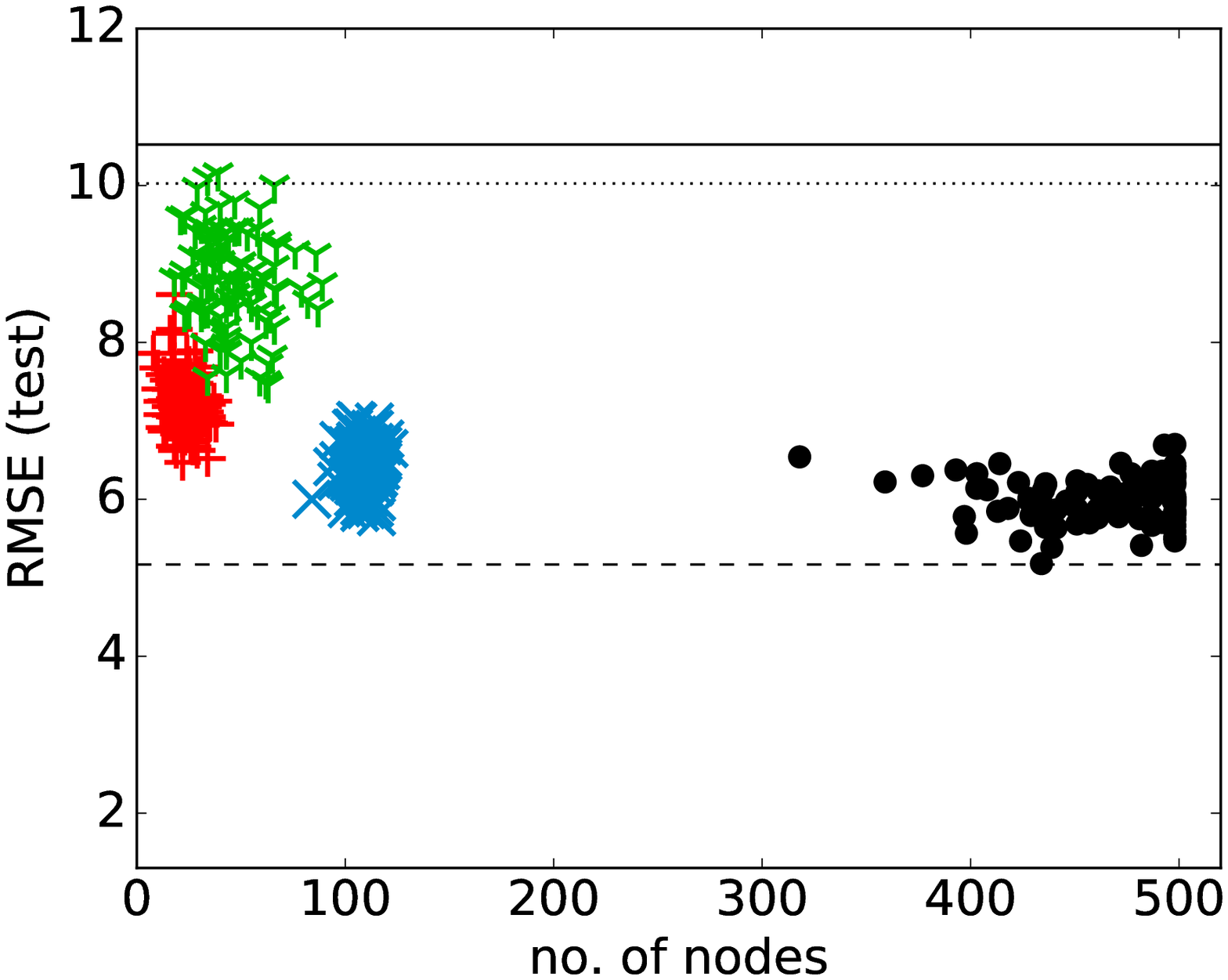}
    \includegraphics[scale=.31, trim=0 0 0 0]{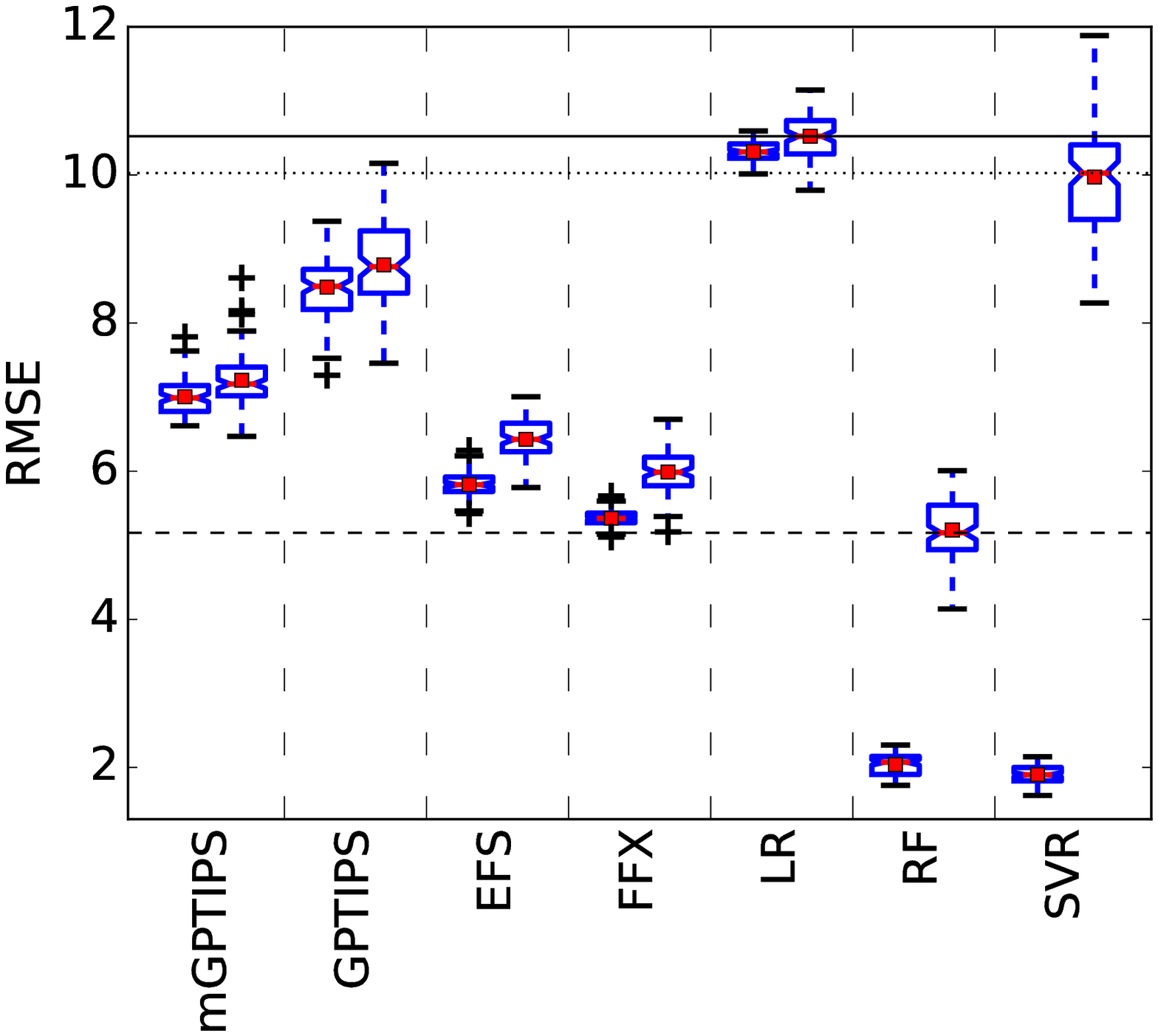}
    \caption{
        Complexity-performance plots (left) and box plots of training and testing errors (right) for the CCS dataset.
        \plotslegend
    }
    \label{fig:ccs}
\end{figure}
\begin{figure}[ht]
    \centering
    \includegraphics[scale=.305, trim=0 -18mm 0 0]{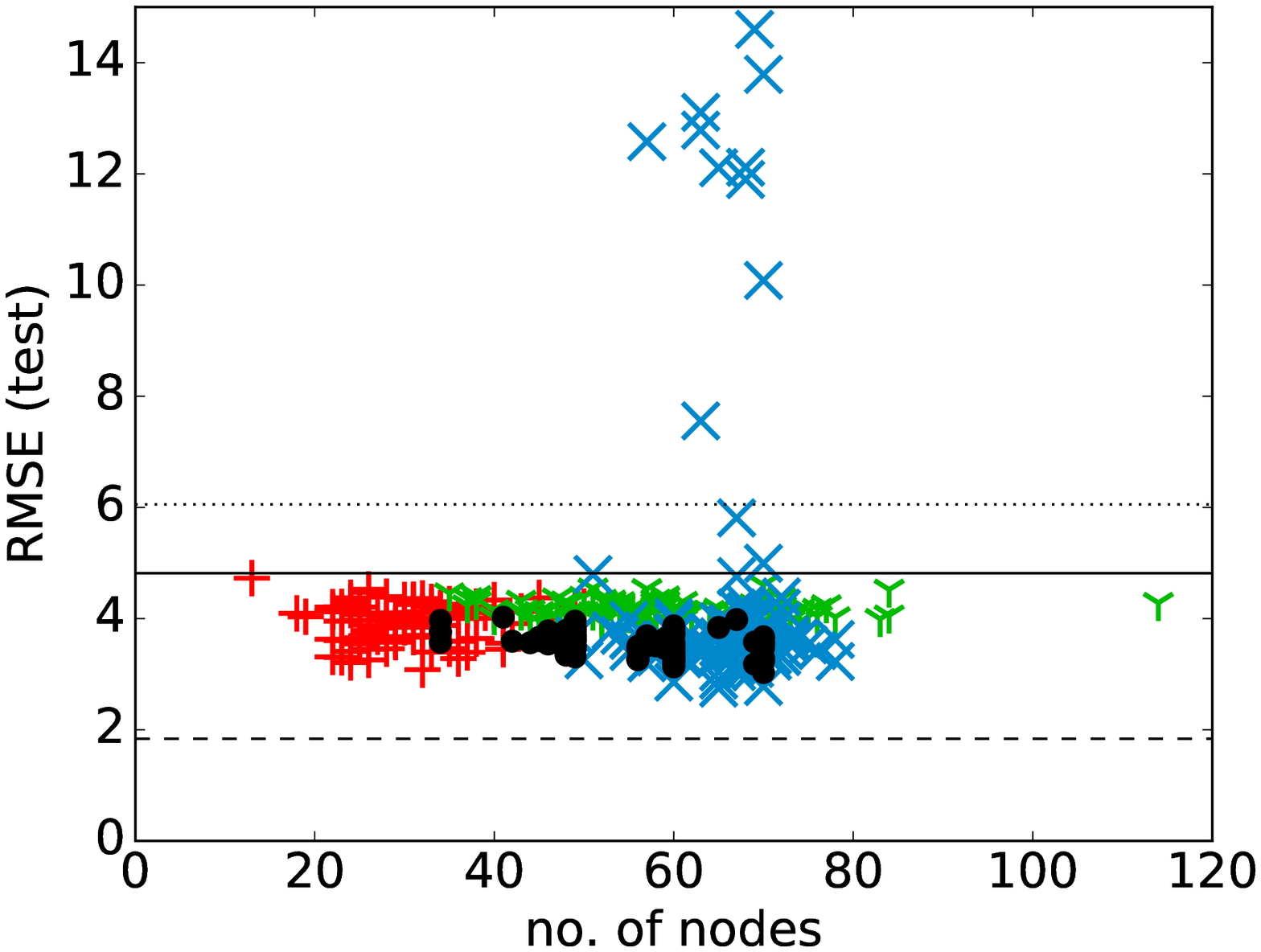}
    \includegraphics[scale=.31, trim=0 0 0 0]{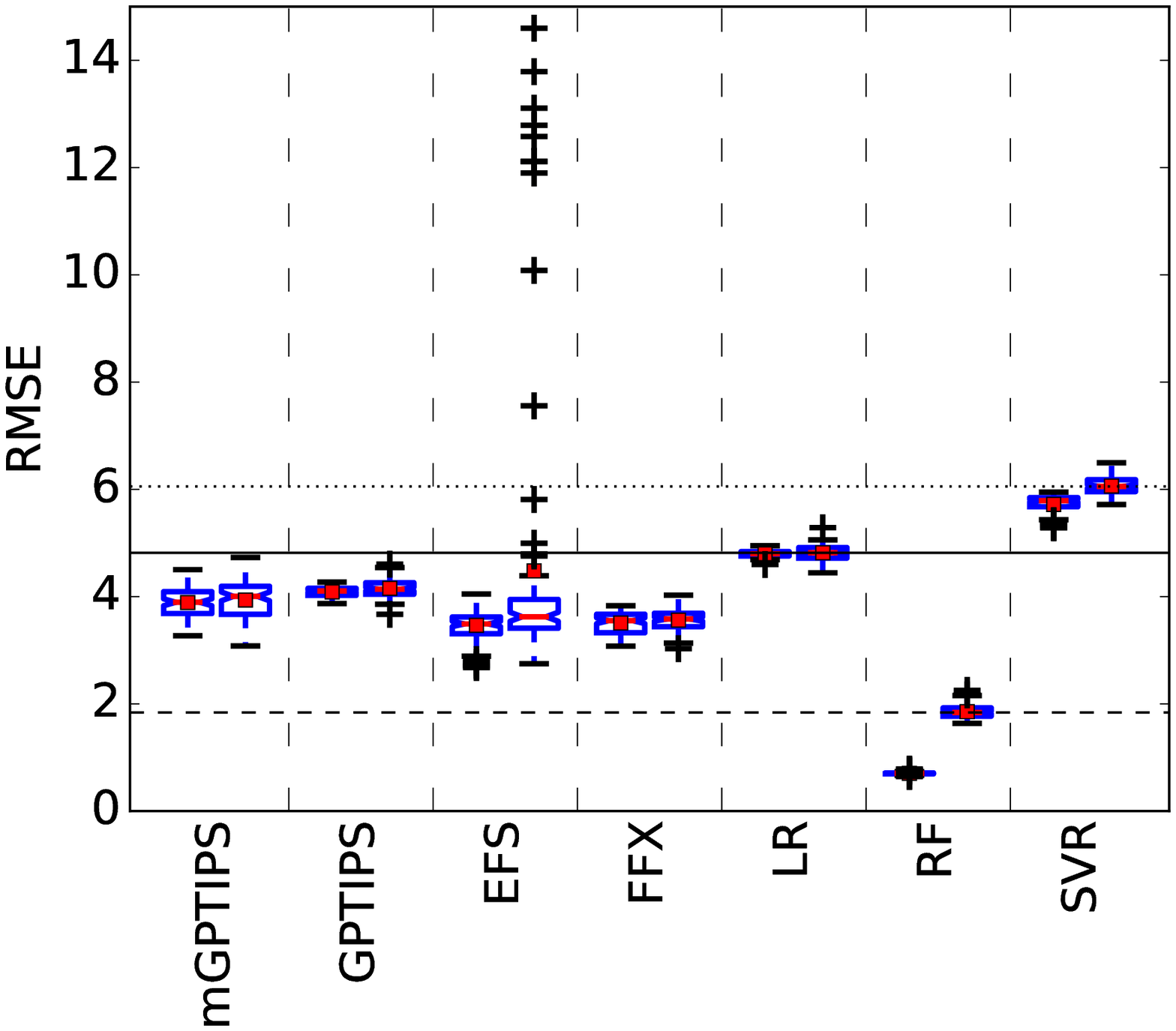}
    \caption{
        Complexity-performance plots (left) and box plots of training and testing errors (right) for the ASN dataset.
        \plotslegend
    }
    \label{fig:asn}
\end{figure}

\runintitle{Koza-1.}
As can be seen from Table \ref{tab:rmses} and Figure \ref{fig:koza1}, GPTIPS was the only method that achieved zero error.
With the default function set it found such model in all runs, although needing more nodes for that.
Enriching the function set (mGPTIPS) enables the method to find simpler models also with optimal performance, but -- due to a larger search space -- it sometimes fails to find the optimum.

FFX and EFS are worse, both reaching RMSE of the order of $10^{-2}$ with no significant difference between them (Table \ref{tab:rmses-stat}), partially due to the large range of RMSE values produced by EFS.
The non-zero error is caused by the regularization used in these methods -- EFS indeed found the optimal bases but their coefficients are not exactly 1.

FFX and GPTIPS tend to construct significantly more complex models (Tables \ref{tab:nodes} and \ref{tab:nodes-stat}) than other methods -- this is most likely caused by the fact they are unable to effectively create the 4th power, and therefore need to compensate for it by creating a lot of bases.

For Koza-1, the SR models are better than or comparable to the tuned ML models.

\runintitle{Korns-11.}
This dataset comes from a quickly changing function with a constant range of values.
The datasets look very much like samples from a constant function with noise.
As can be seen from Tables \ref{tab:rmses}, \ref{tab:rmses-stat} and Figure \ref{fig:korns11}, all the methods (SR and ML) provide models of comparable performance.
The best for this problem is mGPTIPS which is better than the others from the statistical point of view despite the outliers; the real importnace of the difference is, however, questionable.

FFX and mGPTIPS produced significantly simpler models than GPTIPS and EFS (see Tables \ref{tab:nodes} and \ref{tab:nodes-stat}).
Even though FFX is deterministic, the complexity of its models varies highly. 
The only possible cause are the differences in the individual datasets themselves.
Somewhat unexpected is the fact that it influences FFX so much 
compared to the stoachastic EFS. 
Note, however, that despite the larger variance in complexitites, 
the overal complexity of FFX models is still significantly lower than that of EFS models.

\runintitle{S1.}
As can be seen from Table \ref{tab:rmses} and Figure \ref{fig:s1}, the original GPTIPS with the most limited function set among the compared methods, produces complex models with relatively large errors.
FFX produced a simpler model (10 nodes only) with comparable error.
The complexity of EFS models is comparable to FFX, but EFS tends to produce more accurate models.
The best trade-off is provided by mGPTIPS models which are significantly more accurate, with complexities slightly worse than those of EFS.
Note that FFX was run only once since it is a deterministic algorithm and there is only a single instance of this dataset.

The performance of SR models on this benchmark is better than pure linear regression, but worse than RF and SVR.

\runintitle{S2.}
For this problem, the only algorithm that produced models discernibly better than a constant function from a practical point of view was RF.
Out of SR methods, only FFX was able to provide the constant model with only a single node, as can be seen in Table~\ref{tab:nodes} and Figure \ref{fig:s2}.
Default GPTIPS provides models with comparable performance (yet statistically better than FFX), but with much larger complexity.
Some models of mGPTIPS are in fact able to reach better perfomance, but sometimes also much worse (by several orders of magnitude).
EFS provides results similar to mGPTIPS, but more consistent.

\runintitle{UB.}
Except LR, the default GPTIPS is the least accurate solver here, as can be seen in Table \ref{tab:rmses} and Figure \ref{fig:ub}, and also statistically confirmed in Table \ref{tab:rmses-stat}.
Enlarging the function set allows mGPTIPS to find not only more accurate but also simpler models, but still not as good as those provided by the other two SR methods.
The most accurate SR algorithms for this problem are EFS and FFX, with EFS generating models with lower number of nodes than FFX.
Both EFS and FFX, however, produce more complex models than (m)GPTIPS.

Similarly to S1, SR methods are better than pure LR, but worse than SVR and RF.

\runintitle{ENC, ENH.}
As can be seen in Figures \ref{fig:enc} and \ref{fig:enh}, the pattern of the results is similar for both datasets w.r.t. both the accuracy and complexity of the models, which can also be seen in Tables \ref{tab:rmses}-\ref{tab:nodes-stat}.
The results of GPTIPS are dominated both in accuracy and simplicity by mGPTIPS, the results of FFX are dominated by EFS.
EFS and mGPTIPS provide a good compromise with EFS producing more accurate models, while mGPTIPS producing simpler models.

RF and SVR are comparable or better than the best of SR methods, EFS, in terms of accuracy.

\runintitle{CCS.}
In this dataset, a similar pattern among SR algorithms as in ENC and ENH is also present, except that the accuracies of EFS and FFX are flipped, as displayed in Figure \ref{fig:ccs} and Tables \ref{tab:rmses} and \ref{tab:rmses-stat}.

From the complexity point of view, however, the ENC/ENH pattern remains: mGPTIPS provides the simplest models, followed closely by GPTIPS.
EFS produces just over a hundred nodes and, finally, FFX explodes with four to five hundreds of nodes.
The high number of nodes is caused by the majority of bases being the \emph{hinge functions} which carry high complexity

RF models are only slightly, but significantly better than those of the best SR algorithms, FFX and EFS. 
All SR algorithms produce better models than pure linear regression.
Note, however, the failure of SVR on this dataset --- it is better than LR by only a small margin.
Having the best training errors and much worse testing errors, SVR is suspect from overfitting here.

\runintitle{ASN.}
Figure \ref{fig:asn} shows that all of the SR methods perform similarly in terms of RMSE.
From the accuracy point of view (Table \ref{tab:rmses-stat}), EFS and FFX are best (not significantly different from each other), followed by mGPTIPS, and GPTIPS being the worst.
However, EFS, as the only algorithm in this dataset, produced a number of outliers (some actually worse than a pure linear model), and is thus less reliable.

The complexities, however, vary among the algorithms.
The simplest models are produced by mGPTIPS, followed by FFX and GPTIPS which are statistically indifferent (Table \ref{tab:nodes-stat}), and EFS produces the largest models.

RF again produced the most accurate models. 
LR models were in general worse than models of SR methods.
SVR failed again, with both the training and testing errors larger than the errors of LR.
The explanation may lie in the dataset which may be unsuitable for SVR modeling.
Another reason may be the fact that SVR optimizes the hinge loss, and not RMSE.

\subsection{Global Trends}
Across all datasets we can see that none of the compared SR algorithms was the best everywhere, both from the performance and complexity points of view.
We can see that EFS and FFX perform quite well on real-world datasets and the UB artifical dataset, but not as well on the other artificial datasets.
This suggests that for certain class of real-world problems the inability to work with internal constants is not crucial and can be compensated by a linear combination of sufficiently large number of features.

Across all datasets, EFS and FFX methods are very consistent, meaning that the clusters in complexity-performance space are compact and without too many outliers.
This fact might be important in applications where consistency of the produced models is an issue.
In contrast to (m)GPTIPS, this may be the results of the regularized learning employed in EFS and FFX.

(m)GPTIPS tends to have a higher spread of either complexity or accuracy or both (except on Korns-11 where all the algorithms are similarly inconsistent).
We argue that this is caused by the vanilla GP approach based on population of models, in contrast to the population of features of EFS and deterministic generation of features in FFX. 

The comparison of SR methods with conventional ML approaches (with tuned hyperparameters) shows that SR is no silver bullet.
In the majority of cases, the SR approaches were better than pure LR models, but were worse than RF or SVR models.
For many datasets it can also be observed that the differences between training and testing errors were much larger for RF and SVR models, than for SR models.
We thus hypothesize that with the default settings, the SR algorithms were too constrained and produced underfitted models, while the settings found by the grid search for RF and SVR may result in somewhat overfitted models.
If we relaxed the model complexity constraints of the SR algorithms, they may find more accurate models, however the effects on the model interpretability and on the time requirements are not clear and deserve further study.

\subsection{Running Time}
The running times of the methods are presented in Table~\ref{tab:runtimes}.
They are, however, influenced by the implementation language and running environment (FFX runs in Python 2.7, EFS in Java, GPTIPS in MATLAB).
Because of this, the running times are only informative and do not necessarily represent the real complexity of the algorithms.

\begin{table}[ht]
    \centering
    \caption{
        Median running times of the algorithms per dataset (in seconds).
        For RF and SVR, the number in the parentheses denotes the number of points of the grid search.
        The fastest running times among SR algorithms are emphasized.
        }
    \label{tab:runtimes}
    \begin{tabular}{ccccc|ccc}
        \toprule
                 & GPTIPS & mGPTIPS & EFS & FFX & LR & RF (10) & SVR (9) \\
        \midrule
        Koza-1   & 44.98 & 33.51 & {\bfseries 0.33} & 1.71 & $<$0.01 & 2.85 & 0.34 \\
        Korns-11 & 101.91 & 90.18 & 16.38 & {\bfseries 7.86} & $<$0.01 & 71.87 & 1138.74 \\
        S1       & 58.87 & 44.37 & {\bfseries 0.38} & 6.85 & $<$0.01 & 2.88 & 0.54 \\
        S2       & 58.11 & 48.82 & 1.26 & {\bfseries 0.56} & $<$0.01 & 2.97 & 5.71 \\
        UB       & 48.05 & 36.27 & 6.38 & {\bfseries 6.15} & $<$0.01 & 6.05 & 6.67 \\
        \midrule
        ENC      & 57.40 & 51.09 & {\bfseries 24.96} & 109.63 & $<$0.01 & 3.02 & 7.59 \\
        ENH      & 59.63 & 53.23 & {\bfseries 25.42} & 129.83 & $<$0.01 & 2.96 & 11.61 \\
        CCS      & 59.22 & 51.15 & {\bfseries 19.29} & 30.58 & $<$0.01 & 4.73 & 5.32 \\
        ASN      & 68.00 & 57.86 & {\bfseries 9.43} & 22.20 & $<$0.01 & 4.17 & 12.90 \\
        \bottomrule
    \end{tabular}
\end{table}

Based on the wall-clock time, from the SR algorithms, (m)GPTIPS tend to run for the longest time (tens of seconds), with the exception of ENC and ENH datasets, where FFX was even worse.
Runtime of EFS follows the number of features in the dataset: with Koza-1 and S1 (1D) requiring the least time, S2 (2D) requiring a bit more time, followed by Korns-11, UB, and ASN (5D), and finally ENC, ENH, and CCS (8D) requiring the most time.

The time demands of SR methods usually depend only linearly on the number of training examples (since they are used typically only to compute the value of evaluation function).
Conventional ML methods may have much worse dependency on the number of training examples.
This difference is pronounced in our study in case of the SVR algorithm (which needs to compute the kernel matrix) and the Korns-11 benchmark which has a large training set, where SVR is by far the slowest algorithm.
In other cases, the time required to find a symbolic model was more or less comparable to tune and train a conventional ML model (with the exception of pure LR which is of course the fastest among the algorithms).

\section{Conclusions and Future work} \label{sec:conclusion}
In this article we compared three recent methods for symbolic regression, EFS, FFX, and GPTIPS.
All of them produce models from the class of Generalized Linear Models.
Two of those methods, EFS and FFX, use Pathwise Regularized Learning, while GPTIPS uses classical (multiple) linear regression to determine the linear coefficients of the resulting model.
EFS and GPTIPS are stochastic methods based on GP operators of mutation and crossover, while FFX is a completely deterministic method.

We used the methods as off-the-shelf tools, with their default settings and without modifications of their implementations.
Since the default GPTIPS has a very limited function set, we added mGPTIPS with a function set closer to the one used by EFS.

The methods were compared on five artificial and four real world benchmarks.
The results show that none of the algorithms is exceptionally worse or better than the others. 
We have shown some global trends such as the higher tendency of GPTIPS to larger spread in performance.
EFS and FFX turned out to be consistent methods, though not always the best. 

The comparison with tuned conventional ML algorithms shows, that in majority of cases these produced more accurate models, especially RF. 
However, the gap of the SR methods is not large and they produce models with symbolic representation which may be an important asset in circumstances when not only prediction accuracy is important, but also the understanding of the underlying phenomena is required.

\subsection{Future work}
The comparison presented here provides a basic insight into the performance differences between the selected methods.
In the future we plan to expand the set of benchmarks (with varying complexities, higher number of dimensions, and noise), and also expand the set of the compared algorithms, including GSGP.
Another view on the algorithm comparison may be provided by unifying the sets of function symbols of all compared algorithms (which will however require generalizations of some of the presented algorithm implementations).
The expanded set of benchmarks shall also allow us to tune the available parameters of the methods, and thus reduce the effects caused by possibly suboptimal parameter settings.



\begin{acknowledgements}
Jan Žegklitz was supported by the Czech Science Foundation project Nr. \mbox{15-22731S}.
Petr Pošík was supported by the Grant Agency of the Czech Technical University in Prague, grant No. \mbox{SGS14/194/OHK3/3T/13}.
\end{acknowledgements}

\bibliographystyle{spmpsci}      
\bibliography{bibliography}   

\end{document}